\documentclass{article}

\usepackage[main, final]{neurips_2025}

\usepackage[utf8]{inputenc} 
\usepackage[T1]{fontenc}    
\usepackage{hyperref}       
\usepackage{url}            
\usepackage{booktabs}       
\usepackage{amsfonts}       
\usepackage{nicefrac}       
\usepackage{microtype}      
\usepackage{xcolor}         

\usepackage{graphicx}
\usepackage{subcaption}
\usepackage{bm}
\usepackage{algorithm}
\usepackage{algorithmic}
\usepackage{amsmath}
\usepackage{enumitem}
\usepackage{makecell}
\usepackage{multirow}

\usepackage{natbib}
\DeclareMathOperator*{\argmax}{arg\,max}

\title{Centralized Reward Agent for Knowledge Sharing and Transfer in Multi-Task Reinforcement Learning}

\author{
Haozhe Ma$^{1,2}$,
Zhengding Luo$^{3}$\thanks{Corresponding author.},
Thanh Vinh Vo$^{1}$,
Kuankuan Sima$^{4}$,
Tze-Yun Leong$^{1}$ \\
$^{1}$School of Computing, National University of Singapore \\
$^{2}$TikTok Pte. Ltd., Singapore \\
$^{3}$School of Electrical and Electronic Engineering, Nanyang Technological University \\
$^{4}$Department of Electrical and Computer Engineering, National University of Singapore \\
\texttt{\{haozhe.ma, kuankuan\_sima\}@u.nus.edu}, \\ 
\texttt{\{votv, leongty\}@nus.edu.sg}, \texttt{luoz0021@e.ntu.edu.sg}
}

\begin{document}

\maketitle

\begin{abstract}

Reward shaping is effective in addressing the sparse-reward challenge in reinforcement learning (RL) by providing immediate feedback through auxiliary, informative rewards. Based on the reward shaping strategy, we propose a novel multi-task reinforcement learning framework that integrates a centralized reward agent (CRA) and multiple distributed policy agents. The CRA functions as a knowledge pool, aimed at distilling knowledge from various tasks and distributing it to individual policy agents to improve learning efficiency. Specifically, the shaped rewards serve as a straightforward metric for encoding knowledge. This framework not only enhances knowledge sharing across established tasks but also adapts to new tasks by transferring meaningful reward signals. We validate the proposed method on both discrete and continuous domains, including the representative Meta-World benchmark, demonstrating its robustness in multi-task sparse-reward settings and its effective transferability to unseen tasks.

\end{abstract}

\section{Introduction}
\label{sec:introduction}

Reinforcement learning (RL) has made significant progress across various domains, such as robotics~\citep{kober2013reinforcement}, gaming~\citep{lample2017playing}, autonomous vehicles~\citep{aradi2020survey}, signal processing~\citep{luo4837239gfanc}, and large language models~\citep{shinn2024reflexion,ouyang2022training}. However, environments with sparse and delayed rewards remain a significant challenge, as the absence of immediate feedback hinders the agent from distinguishing the value of states and leads to aimless exploration~\citep{expl-survey:ladosz2022exploration}. Reward Shaping (RS) has been proven to be an effective technique for addressing this challenge by providing additional dense and informative rewards~\citep{reward-idea:sorg2010internal,reward-idea:sorg2010reward}. Concurrently, multi-task reinforcement learning (MTRL) is becoming increasingly important due to its ability to transfer knowledge across tasks. In this context, the auxiliary rewards infused with task-specific information in RS offer a straightforward means to distribute knowledge among different tasks. Integrating RS techniques into MTRL is a highly promising and intuitive direction to enhance the efficacy of multi-task learning systems.

Numerous MTRL algorithms for knowledge transfer have been developed. Policy distillation methods identify and combine commonalities across different policies~\citep{rusu2015policy,distral:teh2017distral,parisotto16_actormimic,pmlr-v235-xu24o}; representation sharing methods extract and share the common features or gradients among agents~\citep{yang2020multi,dsharing,sodhani2021multi}; and parameter sharing methods design architectural modules to reuse parameters or layers across networks~\citep{sun2022paco,cheng2023multi}. Despite their potential, these strategies often face slow adaptation to and limited utilization of transferred knowledge. Therefore, leveraging reward shaping, which directly adds a metric to the reward function, offers a compelling alternative to address these limitations.

Regarding reward shaping, not all shaped rewards effectively serve as a medium for knowledge transfer. Specifically, the intrinsic-motivation-based rewards are typically designed using heuristics to generate task-agnostic signals. Examples include incorporating exploration bonuses~\citep{rs-explo:bellemare2016unifying,rs-explo:ostrovski2017count,rs-explo:devidze2022exploration}, rewarding novel states~\citep{rs-explo:tang2017exploration,rs-novelty:burda2018exploration}, and encouraging curiosity-driven behaviors~\citep{rs-curi:pathak2017curiosity,rs-curi:mavor2022stay}. Although these approaches encourage broader exploration, they are not directly related to specific tasks and thus lack transferability. Consequently, we focus on another branch of RS methods, task-contextual rewards, which automatically learn and encode task-specific information, such as hidden values, states contributions, or future-oriented insights, that can be effectively shared across various tasks~\citep{relara:ma2024reward,ma2024highly,mine:durnd,rs-multi:mguni2023learning,memarian2021self}.

To share task-related knowledge in MTRL via RS techniques, and inspired by the \textit{ReLara} framework~\citep{relara:ma2024reward}, which integrates an assistant reward agent to densify sparse environmental rewards, we propose the \textbf{Cen}tralized Reward Agent based MTRL f\textbf{RA}mework (\textbf{CenRA})\footnote{The source code is accessible at: \url{https://github.com/mahaozhe/CenRA}}. The framework consists of two main components: a \textit{centralized reward agent} (CRA) and multiple distributed \textit{policy agents}. Each policy agent individually learns control behaviors within its respective tasks and shares its experiences with the CRA. The CRA extracts common knowledge from these experiences and learns to generate dense rewards that are encoded with task-specific information. These rewards are then distributed back to the policy agents to augment their original environmental rewards. Additionally, given that different tasks may contribute variably to the MTRL system, we introduce an information synchronization mechanism to further balance knowledge distribution by considering task similarity and agent learning progress, thereby ensuring system-wide optimal performance. The main contributions of this paper are summarized as follows:
\vspace{-\parskip}
\begin{enumerate}[label=(\textit{\roman*}),noitemsep,leftmargin=*]
    \item We propose the CenRA framework to address MTRL problems. It incorporates a CRA that functions as a knowledge pool, efficiently distilling and distributing valuable information from various tasks to policy agents while adapting to new tasks.
    \item CenRA leverages reward shaping techniques to infuse insights via dense rewards. This approach not only provides a direct signal for policy agents to absorb knowledge but also effectively addresses the sparse-reward challenge.
    \item We introduce an information synchronization mechanism that considers both task similarity and agent learning progress to balance multi-task learning. This mechanism provides a novel direction for maintaining system equilibrium in MTRL.
    \item CenRA is validated in both discrete and continuous control MTRL environments with sparse extrinsic rewards. CenRA outperforms baseline models in learning efficiency, knowledge transferability, and system-wide performance.
\end{enumerate}

\section{Related Work}

Multi-task reinforcement learning (MTRL) has attracted significant attention recently due to its potential to share knowledge across multiple tasks, thereby improving learning performance~\citep{mtrl:caruana1993multitask}. We discuss existing MTRL literature from three main directions:

\noindent\textbf{Knowledge Transfer} methods focus on identifying and transferring task-relevant features across diverse tasks~\citep{zeng2021decentralized}. Policy distillation~\citep{rusu2015policy} is a well-studied approach to extract and share task-specific behaviors or representations that many works are built on: \cite{distral:teh2017distral} introduced \textit{Distral}, which distills a centroid policy from multiple task-policies; \cite{parisotto16_actormimic} developed \textit{Actor-Mimic}, where a single policy is trained to mimic several expert policies from different tasks; while \cite{yin2017knowledge} incorporated hierarchical prioritized experience replay buffer to select and learn multi-task experiences; \cite{hessel2019multi} further proposed an adaptation mechanism to equalize the impact of each task in policy distillation. Additionally, \cite{xu2020knowledge} explored the transfer of offline knowledge to train policies, and further leveraged online learning for fine-tuning. \cite{picor:bai2023picor} introduced a dual-phase learning approach, optimizing individual policies while correcting them across multiple tasks. \cite{mcal:mysore2022multi} used separate critics for each task to accompany a single actor to integrate their feedback. These methods mitigate gradient interference to an extent, however, balancing the distribution of knowledge across tasks is crucial. Without a careful trade-off, the performance of the entire system could be compromised.

\noindent\textbf{Representation Sharing} methods explore architectural solutions of reusing network modules or representing commonalities to the MTRL problem~\citep{dsharing,devin2017learning,hong2021structure,ma2024mixed,mine:ma2023hierarchical}. \cite{sun2022paco} used a parameter compositional approach to learn and share a subspace of parameters, allowing policies for various tasks to be interpolated within it. \cite{yang2020multi} employed soft modularization to learn foundational policies and utilized a routing network to generate probabilities to combine them. \cite{he2024not} introduced the Dynamic Depth Routing framework, which dynamically adjusts the use of network modules in response to task difficulty. \cite{sodhani2021multi} leveraged task-related metadata to create composable representations. \cite{cheng2023multi} and~\cite{lan2024contrastive} both incorporated attention mechanisms: the former employed attention-based mixture of experts to capture task relationships, while the latter used Temporal Attention for contrastive learning purposes. Although these methods demonstrate efficacy in learning shared representations, they may struggle to fully capture the complexity of highly diverse tasks. Moreover, adapting shared structures to new tasks typically requires extra design efforts.

\noindent\textbf{Single-Policy Generalization} methods learn a single policy to solve multiple tasks simultaneously or continuously, in the absence of information from prior policies or task-specific details, in which case, the primary goal is to enhance the policy's generalization capabilities. Model-free meta-learning techniques have been proposed to enhance the multi-task generalization~\citep{finn2017model}. \cite{yang2017multi} designed a sharing network structure that allows an agent to learn multiple tasks concurrently. \cite{vuong2019sharing} introduced a confidence-sharing agent to detect and define shared regions between tasks to support single policy learning. \cite{wan2020mutual} proposed a transfer learning framework to handle mismatches in state and action spaces. Additionally, several methods focus on overcoming gradient interference to enhance the generalization in various tasks~\citep{chen2018gradnorm,yu2020gradient}, while \cite{ammar2014online} developed a consecutive learning policy gradient approach. These methods are efficient in saving computational resources, but the generalization ability of the policy may be constrained when faced with out-of-distribution or previously unseen tasks.

\section{Preliminaries}

\noindent\textbf{Markov Decision Process (MDP)} models sequential decision-making problems under uncertainty. An MDP represents the interaction between an agent and its environment as a tuple $\langle S, A, P, R, \gamma \rangle$, where $S$ is the state space, $A$ is the action space, $P: S \times A \times S \rightarrow [0,1]$ is the probability of transiting from one state to another given an action, $R: S \times A \rightarrow \mathbb{R}$ is the reward function, and $\gamma \in [0,1]$ is the discount factor to modulate the importance of future versus immediate reward. 

\noindent\textbf{Multi-Task Reinforcement Learning (MTRL)} addresses the challenge of learning multiple tasks simultaneously within an integrated model to leverage commonalities and differences across tasks. Typically, MTRL introduces a task space $\mathcal{T}$, assuming all tasks are sampled from this space and thus follow a unique distribution. Each task is modeled as an independent MDP. An MTRL agent aims to learn optimal policies $\pi_i: S \rightarrow A$ for each task $T_i \sim \mathcal{T}$, to maximize their corresponding expected cumulative rewards, or returns, denoted by $G_i=\mathbb{E}[\sum_{t=0}^{\infty} \gamma^t R_i(s_t, a_t)]$. 

\noindent\textbf{RL with an Assistant Reward Agent (ReLara)}~\citep{relara:ma2024reward} introduces a dual-agent framework designed to tackle the challenge of sparse rewards in RL. Within this framework, the original agent is termed as \textit{policy agent}, while an assistant \textit{reward agent} is integrated to enrich the feedback mechanism by generating dense, informative rewards. The reward agent, trained as a self-contained RL agent, autonomously extracts hidden value information from the environmental states and the actions of the policy agent to craft meaningful reward signals. These signals significantly improve learning efficiency by providing immediate and pertinent bonuses.

\section{Methodology}

\begin{figure*}[t]
    \centering
    \includegraphics[width=\textwidth]{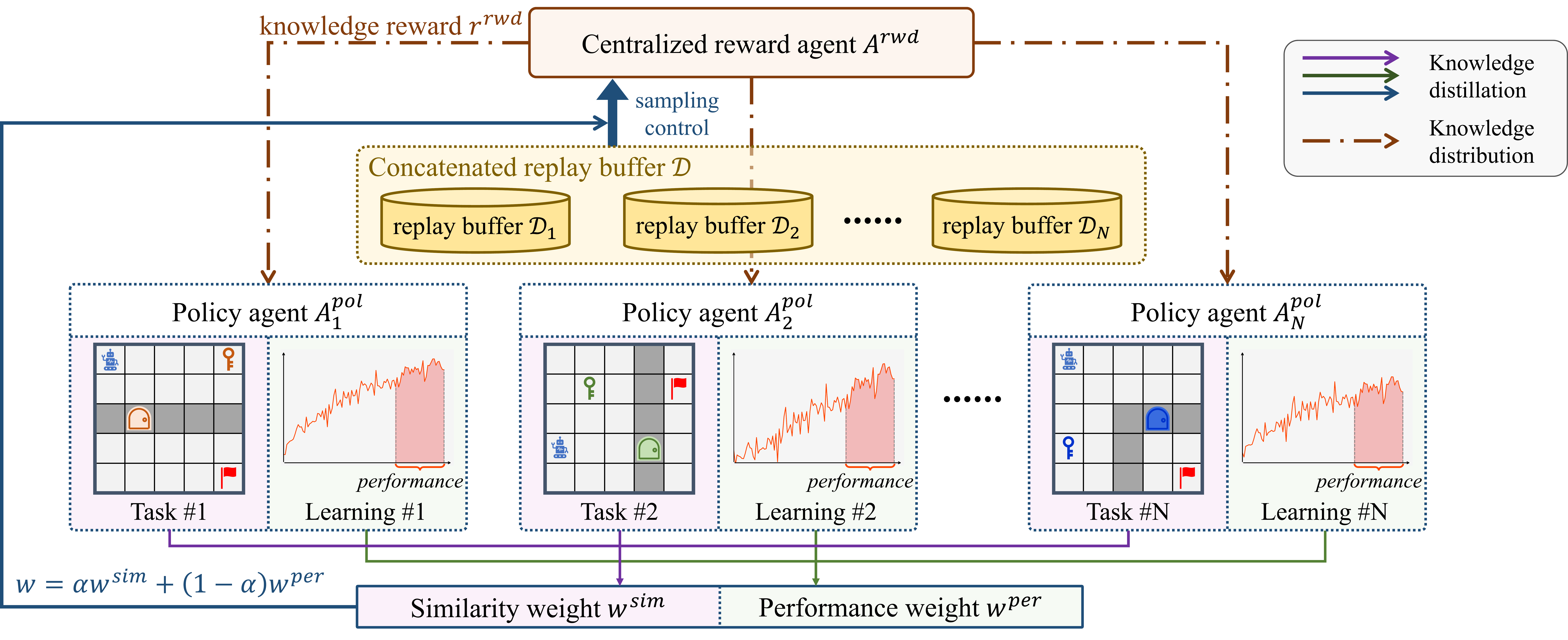}
    \caption{A high-level illustration of the CenRA framework. The centralized reward agent functions as a knowledge repository, distilling information from various tasks and distributing it to individual policy agents to enhance learning efficiency.}
    \label{fig:framework}
\end{figure*}

We propose the \textbf{Cen}tralized Reward Agent f\textbf{RA}mework (\textbf{CenRA}) for MTRL, which incorporates a \textit{centralized reward agent} (CRA) to support multiple reinforcement learning agents across multiple tasks. A high-level illustration of the CenRA framework is shown in Figure~\ref{fig:framework}. The CRA is responsible for extracting general task-specific knowledge from various tasks and distributing valuable information to the policy agents by reconstructing their reward models. The detailed methodology for knowledge extraction and sharing is presented in Section~\ref{sec:met-share}. Furthermore, to mitigate the potential disparities in the information that each task contributes, which might lead to an imbalance in knowledge distribution, we introduce an information synchronization mechanism by considering two main factors: the similarity of the tasks and the online learning performance of the policy agents, details given in Section~\ref{sec:met-sync}. Finally, the overall framework of CenRA is presented in Section~\ref{sec:met-framework}.

\subsection{Knowledge Distillation and Distribution}
\label{sec:met-share}

\subsubsection{Problem Formulation}

We consider an MTRL setting comprising $N$ distinct tasks $\{T_1, T_2, \dots, T_N\}$, all executed within the same type of environment $\mathcal{E}$. We assume that the shape of state $s \in S$ and action $a \in A$ remain uniform across tasks, to ensure the CRA processes consistent inputs. Despite this uniformity, each task may feature different state spaces, action spaces, goals, and transition dynamics. For instance, a series of mazes with the same size but varying map configurations would satisfy this condition. For each task $T_i$, we denote the transition function as $P_i(s' | s, a)$ and the reward function as $R_i(s, a)$.

The centralized reward agent (CRA) is denoted as $\mathcal{A}^{rwd}$ and multiple policy agents are denoted as $\{\mathcal{A}^{pol}_1,\allowbreak \mathcal{A}^{pol}_2, \dots, \mathcal{A}^{pol}_N\}$. Each policy agent $\mathcal{A}^{pol}_i$ operates independently to complete its corresponding task $T_i$, utilizing appropriate RL algorithms as backbones. For example, implementing DQN~\citep{dqn:mnih2015human} for discrete control tasks, while TD3~\citep{fujimoto2018addressing} or SAC~\citep{sac:haarnoja2018soft} for continuous control tasks. Moreover, the policy of CRA $\mathcal{A}^{rwd}$ is $\pi^{rwd}$, and the internal policy of policy agent $\mathcal{A}^{pol}_i$ is $\pi^{pol}_i$.

\subsubsection{Centralized Reward Agent}

The CRA $\mathcal{A}^{rwd}$ aims to extract environment-relevant knowledge and distribute it to policy agents by generating additional dense rewards to support their original reward functions. Similar to the ReLara framework~\citep{relara:ma2024reward}, we model the CRA as a self-contained RL agent, yet, as an extension to ReLara, our CRA is designed to concurrently interact with multiple policy agents and their respective tasks. The CRA's policy $\pi^{rwd}$ generates continuous rewards given both an environmental state and a policy agent's behavior. Specifically, $\pi^{rwd}$ maps the Cartesian product of the state space and action space, $S \times A$, to a defined \textit{reward space}, which constrains the rewards to a range of real numbers, $\mathcal{R} = [R_{min}, R_{max}] \subset \mathbb{R}$. For simplicity, we denote the observation of the CRA as $s^{rwd}=(s_i,a_i)$, where $s_i \sim T_i$ and $a_i \sim \pi^{pol}_i(s_i)$. To distinguish from the environmental reward, the generated reward is termed as \textit{knowledge reward}, denoted as $r^{knw}$.

We adopt an off-policy actor-critic algorithm to optimize the CRA~\citep{konda1999actor}. To aggregate and reuse experiences from all policy agents, a concatenated replay buffer $\mathcal{D} = \bigcup_{i=1}^N{\mathcal{D}_i}$ is constructed, where $\mathcal{D}_i$ represents the replay buffer of each policy agent $\mathcal{A}^{pol}_i$. Besides, each transition is augmented with the CRA-generated knowledge reward, $r^{knw}$. Specifically, the transition from policy agent $\mathcal{A}^{pol}_i$ stored in the replay buffer is defined as $\tau = (s^{rwd}_t, r^{knw}_t, r^{env}_t, s^{rwd}_{t+1} | T_i)$. The augmented transition includes all necessary information for optimizing both the CRA and each corresponding policy agent, thus making the concatenated replay buffer a shared resource across the entire framework and minimizing storage overhead.

The CRA's update process involves using these stored transitions to optimize the reward-generating actor $\pi^{rwd}$ and the value estimation critic. The objective function for the critic module is:
\begin{equation}
    J(V^{rwd}) = \underset{\tau_t \sim \mathcal{D}}{\mathbb{E}}[\delta_t^2], \quad \delta_t = r^{env}_t+\gamma V^{rwd}(s^{rwd}_{t+1})-V^{rwd}(s^{rwd}_t) | T_i,
\end{equation}
where $\tau_t = (s^{rwd}_t,r^{env}_t,s^{rwd}_{t+1} | T_i) \sim \mathcal{D}$. Concurrently, the actor module is updated through the following objective function:
\begin{equation}
    J(\pi^{rwd}) = \underset{\tau_t \sim \mathcal{D}}{\mathbb{E}} \Big[\underset{r^{knw}_t \sim \pi^{rwd}(\cdot | s^{rwd}_t)}{\mathbb{E}} \big[\log \pi^{rwd}(r^{knw}_t | s^{rwd}_t) \cdot \delta_t \big]\Big].
\end{equation}

\subsubsection{Policy Agents with Knowledge Rewards}

Each policy agent $\mathcal{A}^{pol}_i$ stores the experiences in its corresponding replay buffer $\mathcal{D}_i$. They receive two types of rewards: the environmental reward $r^{env}_i$ from their respective task $T_i$ and the knowledge reward $r^{knw}$ from CRA. The augmented reward is given by: 
\begin{equation}
    r^{pol}_i = r^{env}_i + \lambda r^{knw}, \quad r^{knw} \sim \pi^{rwd}(\cdot | s_i, a_i),
\end{equation}
where $\lambda \in (0,1]$ is a scaling weight factor. The optimal policy ${\pi^{pol}_i}^*$ for each agent is derived by maximizing the cumulative augmented reward:
\begin{equation}
    {\pi^{pol}_i}^* = \argmax_{\pi^{pol}_i} \underset{(s_i, a_i) \sim \pi^{pol}_i}{\mathbb{E}} \Big[ \sum_{t=0}^{\infty} \gamma^t r^{pol}_i \Big].
\end{equation}
It is worth noting that the environmental reward $r^{env}_i$ is retrieved from the replay buffer (if adopting an off-policy approach). Conversely, the knowledge reward $r^{knw}$ is computed in real-time using the most recently updated $\mathcal{A}^{rwd}$, ensuring it reflects the latest learning advancements. Lastly, each policy agent is able to employ any suitable RL algorithm, whether on-policy or off-policy, to best address its specific task, which enhances the CenRA framework's generality and flexibility.

\subsection{Information Synchronization of Policy Agents}
\label{sec:met-sync}

In the CenRA, the information provided by different tasks may exhibit significant disparities, potentially leading to an imbalance in knowledge extraction and distribution. We introduce an information synchronization mechanism for CenRA to maintain a balanced manner from the perspective of the entire system. Specifically, we control the quantity of samples that CRA retrieves from each task's replay buffer $\mathcal{D}_i$ by a \textit{sampling weight} $\bm{w}$, by considering two aspects: the similarity among tasks and the real-time learning performance of the policy agents.

\noindent\textbf{Similarity Weight} $\bm{w}^{sim}$ is derived from the similarity among tasks, enabling the CRA to focus on relatively outlier tasks. To simplify computation, we use the hidden layers extracted from each policy agent's neural network encoders to represent the tasks' features. To reduce randomness, we average the hidden features of the most recent $K$ steps. We adopt a cross-attention mechanism to calculate the similarity weight~\citep{vaswani2017attention}. Specifically, for task $T_i$, let $\bm{H}_i$ denote the averaged hidden feature vector, which serves as the \textit{key}, and the centroid of all tasks $\bm{c}$ acts as the \textit{query}. Then, the similarity $s_i$ of task $T_i$ to the centroid of the task cluster is calculated as:
\begin{equation}
    s_i = \frac{\bm{c}^T \cdot \bm{H}_i}{\sqrt{D}}, \quad \bm{c} = \frac{1}{N}\sum_{k=1}^{N}\bm{H}_k,
\end{equation}
where $D$ is the dimension of the hidden feature to prevent gradient vanishing or exploding. A larger $s_i$ indicates a greater similarity between $T_i$ and the centroid. It is worth noting that, to avoid the centroid $\bm{c}$ approaching zero due to feature vectors $\bm{H}_i$ pointing in opposite directions, all latent representations $\bm{H}_i$ in our framework are extracted from ReLU activation layers. This ensures that every element of $\bm{H}_i$ is non-negative, effectively preventing feature cancellation and maintaining a well-defined and numerically stable centroid $\bm{c}$. Given our assumption is that the tasks farther from the centroid require more attention, the \textit{similarity weight} is defined as $\bm{w}^{sim} = \text{Softmax}\big([1/s_1, 1/s_2, \dots, 1/s_N]\big)$. 

\noindent\textbf{Performance Weight} is determined by the real-time learning performance of each policy agent, to ensure the CRA focuses more on lagging tasks. Similar to the similarity weight, we average the environmental rewards $r^{env}_i$ from the most recent $K$ steps, denoted as $R^{tail}_i$, to measure the recent learning trends. The \textit{performance weight} is then defined as $\bm{w}^{per} = \text{Softmax}\big([1/R^{tail}_1, 1/R^{tail}_2, \dots, 1/R^{tail}_N]\big)$.

The final \textit{sampling weight} $\bm{w}$ is formulated as $\bm{w} = \alpha \bm{w}^{sim} + (1-\alpha) \bm{w}^{per}$, where $\alpha$ is a hyperparameter to balance the two aspects. The CRA samples from each replay buffer $\mathcal{D}_i$ according to $\bm{w}$, ensuring a balanced and effective knowledge extraction and learning.

\subsection{Overall Framework}
\label{sec:met-framework}

The overall framework of CenRA is summarized in Algorithm~\ref{alg:main-algo}. The CRA and policy agents are updated alternately and asynchronously, with the frequency of updating the CRA adjustable according to the actual situation. Sampling weights are calculated in real-time, using the most recently optimized encoders and the current learning performance, ensuring CRA continuously adjusts its focus to optimally balance knowledge extraction across multiple tasks.

The learned CRA acts as a robust knowledge pool, which is able to support new tasks by transferring knowledge through auxiliary reward signals. This is particularly beneficial in sparse-reward environments, as the knowledge rewards can guide the policy agents toward the correct direction and reduce exploration burden. Additionally, the CRA can be further optimized alongside new tasks in a continuous learning scheme that enhances adaptability and effectiveness in dynamic settings.

\begin{algorithm}[h]
\caption{Centralized Reward Agent based MTRL}
\label{alg:main-algo}
\begin{algorithmic}[1]
    \REQUIRE Multiple tasks $\{T_1, T_2, \dots, T_N\}$.
    \REQUIRE Policy agents $\{\mathcal{A}^{pol}_1, \mathcal{A}^{pol}_2, \dots, \mathcal{A}^{pol}_N\}$.
    \REQUIRE Centralized reward agent $\mathcal{A}^{rwd}$.
    \REQUIRE Concatenated replay buffer $\mathcal{D} = \bigcup_{i=1}^N{\mathcal{D}_i}$.
    
    \FOR{each iteration}
        \FOR{each task $T_i$}
        \STATE $(s_t, a_t, r^{env}_t, s_{t+1}, a_{t+1}) \sim \text{Interact}(\mathcal{A}^{pol}_i, T_i)$  \hfill $\triangleright$ Interact and collect one transition
        \STATE $r^{knw}_t \sim \mathcal{A}^{rwd}(s_t, a_t)$ \hfill $\triangleright$ Sample an off-policy knowledge reward
        \STATE $s^{rwd}_t = (s_t, a_t)$, $s^{rwd}_{t+1} = (s_{t+1}, a_{t+1})$
        \STATE $\mathcal{D}_i \leftarrow \mathcal{D}_i \cup \{(s^{rwd}_t, r^{knw}_t, r^{env}_t, s^{rwd}_{t+1} | T_i)\}$  \hfill $\triangleright$ Store the transition in corresponding $\mathcal{D}_i$
        \STATE Update policy agent $\mathcal{A}^{pol}_i$  \hfill $\triangleright$ Update $\mathcal{A}^{pol}_i$ using backbone RL algorithm
        \ENDFOR
        \STATE $\bm{w} = \alpha \bm{w}^{sim} + (1-\alpha) \bm{w}^{per}$ \hfill $\triangleright$ Calculate sampling weight
        \STATE $\{s^{rwd}_t,r^{knw}_t,r^{env}_t,s^{rwd}_{t+1} | T_i\}_{\mathcal{B}} \sim \mathcal{D} | \bm{w}$ \hfill $\triangleright$ Draw samples based on the sampling weight
        \STATE Update centralized reward agent $\mathcal{A}^{rwd}$
    \ENDFOR
\end{algorithmic}
\end{algorithm}

\section{Experiments}

\begin{figure}[h]
\centering
\begin{subfigure}[b]{0.54\textwidth}
    \centering
    \includegraphics[width=\textwidth]{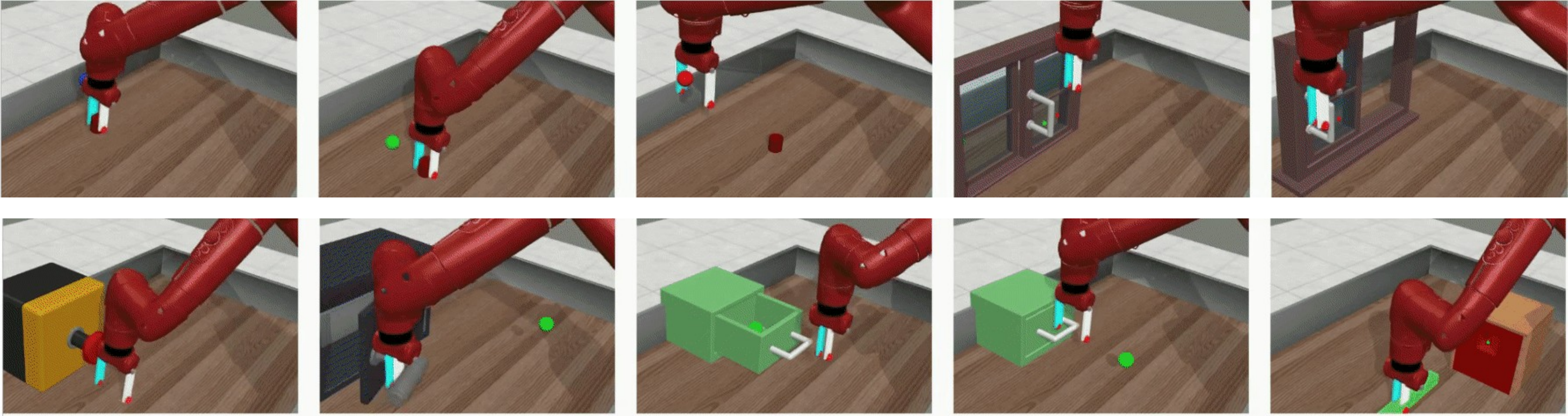}
    \caption{\textit{Meta-World (ML10-sparse and ML50-sparse)}}
\end{subfigure}
\hfill
\begin{subfigure}[b]{0.145\textwidth}
    \centering
    \includegraphics[width=\textwidth]{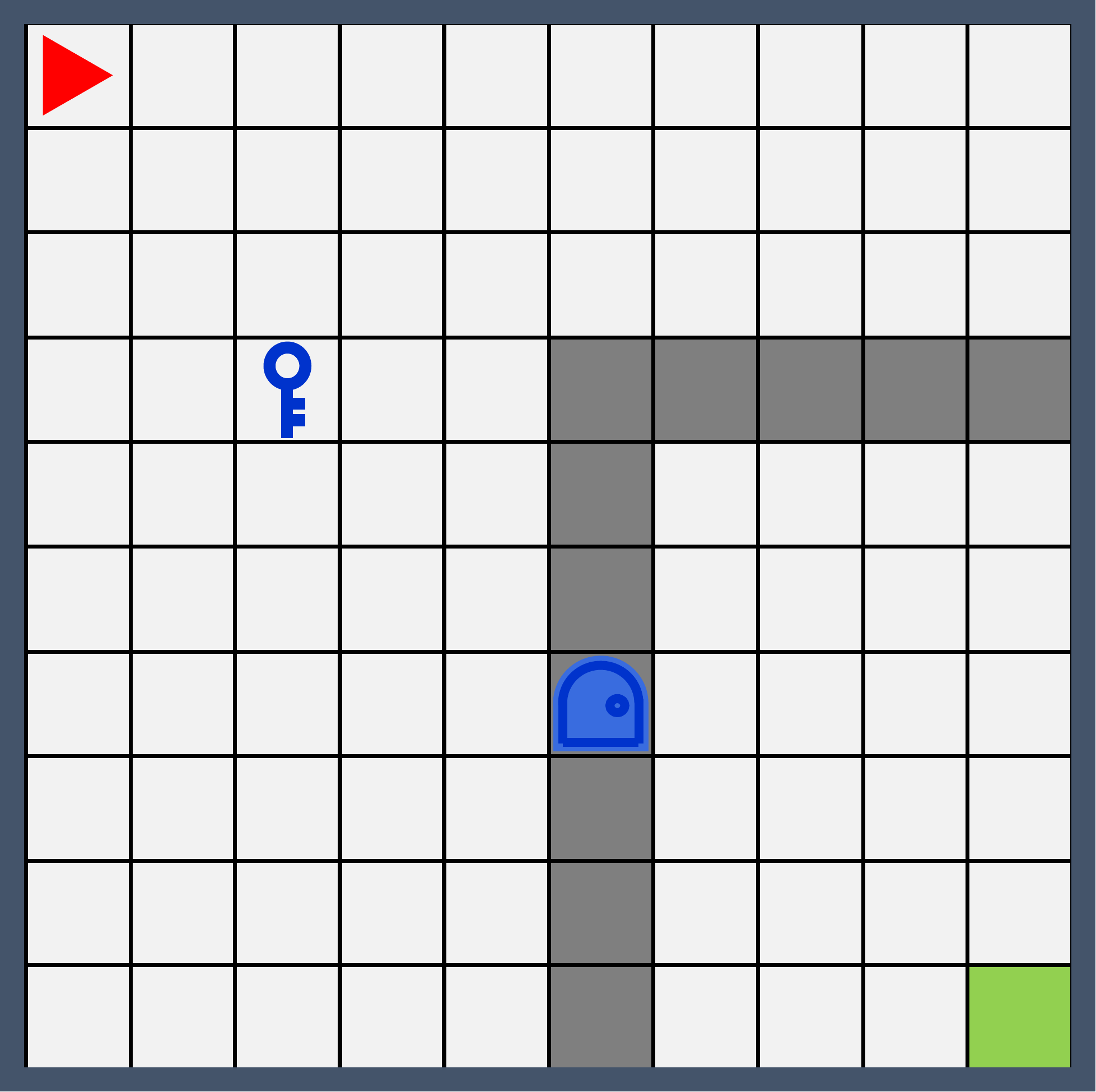}
    \caption{\textit{2DMaze}}
\end{subfigure}
\hfill
\begin{subfigure}[b]{0.145\textwidth}
    \centering
    \includegraphics[width=\textwidth]{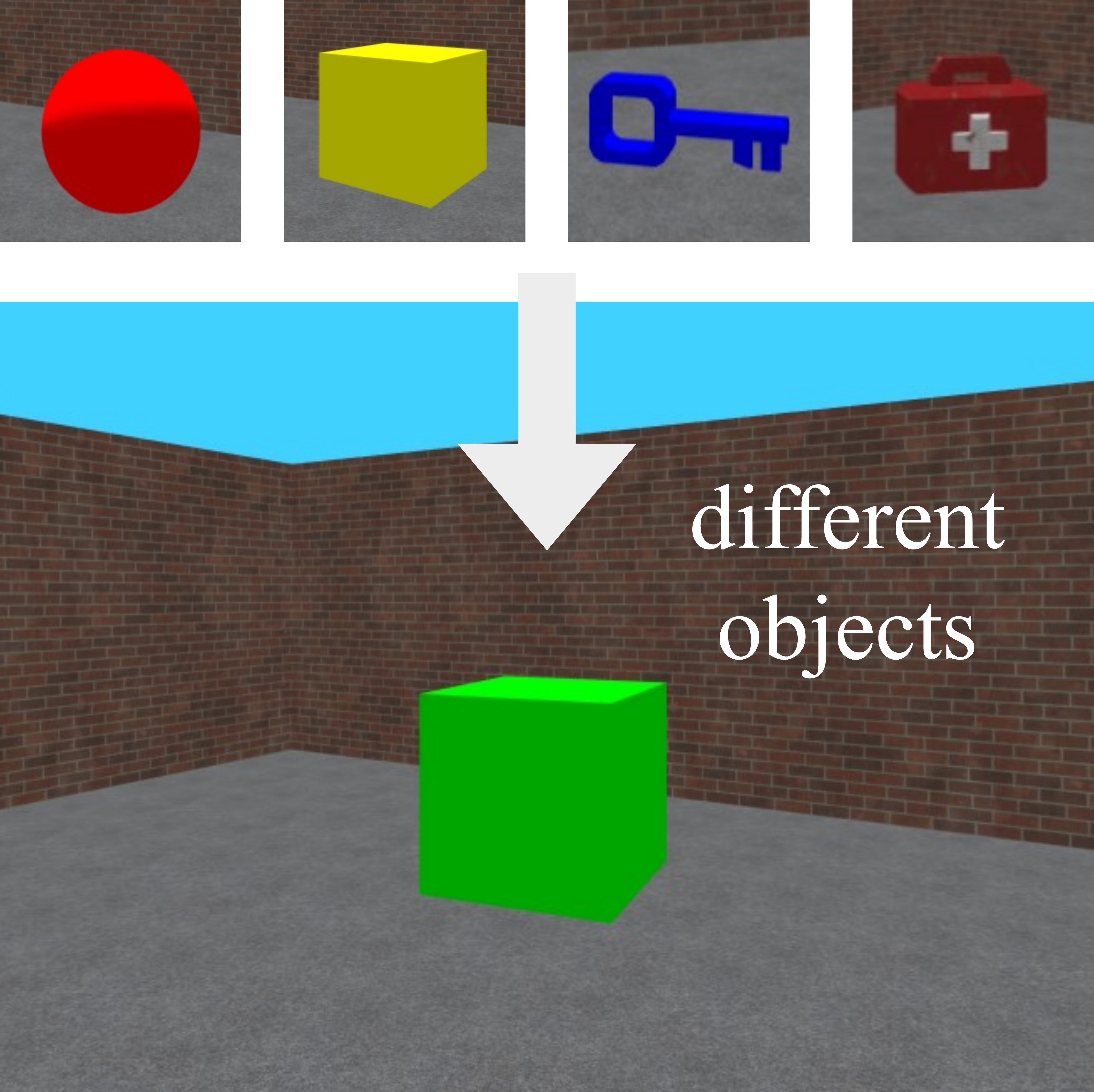}
    \caption{\textit{3DPickup}}
\end{subfigure}
\hfill
\begin{subfigure}[b]{0.145\textwidth}
    \centering
    \includegraphics[width=\textwidth]{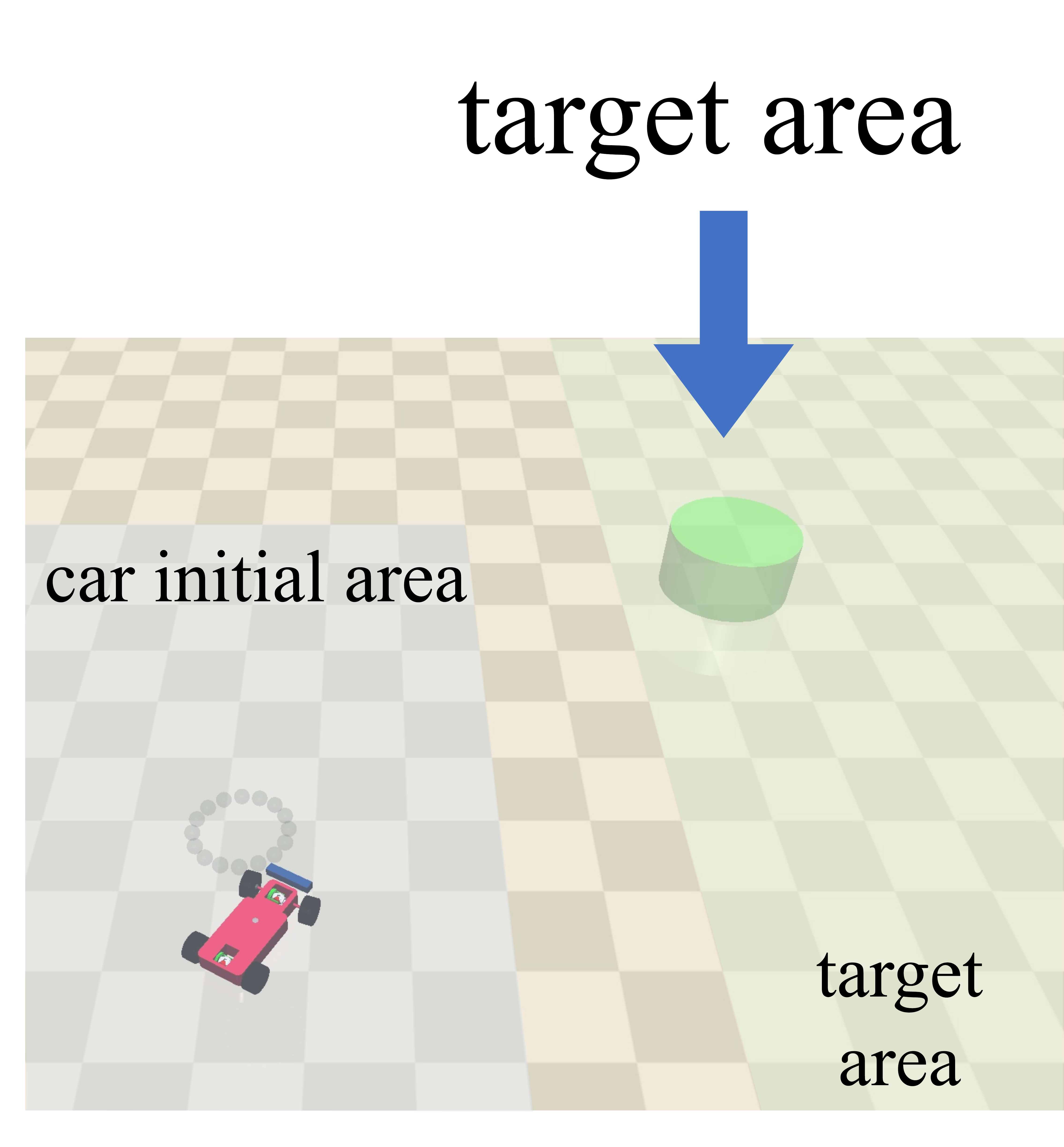}
    \caption{\textit{MujocoCar}}
\end{subfigure}
\caption{Environments with multiple tasks. (a) \textbf{Meta-World}: two sparse-reward versions are used: \textbf{ML10-sparse} and \textbf{ML50-sparse}, including diverse robotic manipulation tasks. (b) \textbf{2DMaze}: 2D maze tasks where the agent must pick up a key and then pass through a door to exit. (c) \textbf{3DPickup}: 3D maze tasks where the agent aims to navigate to and pick up different target objects at different locations. (d) \textbf{MujocoCar}: mujoco-based race car aims to navigate to different specified areas.}
\label{fig:tasks}
\end{figure}

We conduct experiments in four MTRL domains as shown in Figure~\ref{fig:tasks}: the widely used \textbf{Meta-World} benchmark (including \textbf{ML10} with 10 tasks and \textbf{ML50} with 50 tasks)~\citep{env-mw:yu2020meta}, \textbf{2DMaze}, \textbf{3DPickup}~\citep{env-mini:chevalier2024minigrid}, and \textbf{MujocoCar}~\citep{env-car:ji2023safety}. All tasks, including those in Meta-World, are crafted to provide \textbf{sparse environmental rewards}, where the agent receives a reward of $1$ only upon successful completion of the final objective, and $0$ otherwise. The detailed task configurations are provided in Appendix~\ref{app:tasks}. 

\subsection{Comparative Evaluation in MTRL}
\label{sec:comp}

We benchmark CenRA against several state-of-the-art baselines: (a) the backbone RL algorithms of the policy agents: DQN~\citep{dqn:mnih2015human} for discrete control tasks and SAC~\citep{sac-app:haarnoja2018soft} for continuous control tasks; (b) the ReLara algorithm~\citep{relara:ma2024reward}, which can be regarded as a decentralized variant of CenRA, where each policy agent is paired with a separate reward agent, without cross-task information sharing; (c) the TD-MPC2 algorithm~\citep{TD-MPC2:hansen2024td}; (d) the Contrastive Modules with Temporal Attention (CMTA) algorithm~\citep{lan2024contrastive}; (e) the Policy Optimization and Policy Correction (PiCor) algorithm~\citep{picor:bai2023picor}; (f) the Multi-Critic Actor Learning (MCAL) algorithm~\citep{mcal:mysore2022multi}; (g) the Parameter-compositional MTRL (PaCo) algorithm~\citep{sun2022paco}; (h) the Shared-Critic (SC) algorithm~\citep{sc:zhang2021multi}; and (i) the MTRL with Soft Modularization (SoftModule)~\citep{yang2020multi}. They are implemented by either the \textit{CleanRL} library~\citep{cleanrl:huang2022cleanrl} or official codebases. Each task is trained with 10 different random seeds, and the average results are reported.

In the \textit{Meta-World} domain, \textit{ML10-sparse} provides 10 training tasks and 5 held-out test tasks, while \textit{ML50-sparse} includes 45 training tasks and 5 test tasks. For the remaining domains, each consists of 4 training tasks and 1 test task. In this section, we evaluate the final returns achieved by the trained agents, averaged over all training tasks in each domain, as shown in Table~\ref{tab:comparison-returns}. We additionally report the episodic returns and their standard errors throughout training in the \textit{2DMaze}, \textit{3DPickup}, and \textit{MujocoCar} domains in Figure~\ref{fig:comparison}. To ensure a fair comparison, we adopt consistent hyperparameters (where applicable) and identical network architectures across all experiments; detailed configurations are provided in Appendix~\ref{app:configs}.

\begin{figure*}[t]
    \centering
    \includegraphics[width=\textwidth]{Images/comp-all.pdf}
    \caption{Comparison of CenRA with baselines in \textit{2DMaze}, \textit{3DPickup}, and \textit{MujocoCar} domains.}
    \label{fig:comparison}
\end{figure*}

\begin{table}[t]
\centering
\caption{Episodic returns (mean $\pm$ standard error) of all trained agents tested over 100 episodes and averaged across all training tasks in each domain ($\uparrow$ higher is better).}
\label{tab:comparison-returns}
\small

\begin{tabular}{cccccc}
\toprule
Algorithm & \textit{ML10-sparse} & \textit{ML50-sparse} & \textit{2DMaze} & \textit{3DPickup} & \textit{MujocoCar} \\
\midrule
CenRA (ours) & \textbf{0.875 $\pm$ 0.121} & \textbf{0.755 $\pm$ 0.034} & \textbf{0.913 $\pm$ 0.023} & \textbf{0.880 $\pm$ 0.060} & \textbf{514.875 $\pm$ 0.675} \\
DQN/SAC      & 0.256 $\pm$ 0.056 & 0.189 $\pm$ 0.012 & 0.645 $\pm$ 0.070 & 0.243 $\pm$ 0.048 & 198.000 $\pm$ 0.453 \\
ReLara       & 0.674 $\pm$ 0.105 & 0.541 $\pm$ 0.057 & 0.803 $\pm$ 0.065 & 0.565 $\pm$ 0.088 & 429.800 $\pm$ 0.655 \\
TD-MPC2      & 0.823 $\pm$ 0.091 & 0.608 $\pm$ 0.032 & 0.884 $\pm$ 0.046 & 0.712 $\pm$ 0.051 & 505.341 $\pm$ 0.712 \\
CMTA         & 0.787 $\pm$ 0.076 & 0.603 $\pm$ 0.026 & 0.753 $\pm$ 0.037 & 0.695 $\pm$ 0.043 & 480.187 $\pm$ 0.623 \\
PiCor        & 0.865 $\pm$ 0.230 & 0.672 $\pm$ 0.123 & 0.818 $\pm$ 0.053 & 0.438 $\pm$ 0.085 & 437.550 $\pm$ 0.663 \\
MCAL         & 0.842 $\pm$ 0.067 & 0.605 $\pm$ 0.055 & 0.885 $\pm$ 0.080 & 0.548 $\pm$ 0.068 & 369.200 $\pm$ 0.595 \\
PaCo         & 0.854 $\pm$ 0.045 & 0.582 $\pm$ 0.022 & 0.834 $\pm$ 0.057 & 0.557 $\pm$ 0.072 & 421.210 $\pm$ 0.635 \\
SC           & 0.556 $\pm$ 0.063 & 0.354 $\pm$ 0.023 & 0.798 $\pm$ 0.052 & 0.687 $\pm$ 0.038 & 400.254 $\pm$ 0.518 \\
SoftModule   & 0.630 $\pm$ 0.042 & 0.423 $\pm$ 0.057 & 0.822 $\pm$ 0.076 & 0.486 $\pm$ 0.055 & 355.125 $\pm$ 0.594 \\
\bottomrule
\end{tabular}
\end{table}

We observe that CenRA consistently outperforms all baselines in three main aspects. First, it achieves the highest episodic returns in all tasks, demonstrating superior learning efficiency and faster convergence. Moreover, it demonstrates good stability and robustness, exhibiting fewer fluctuations and oscillations, especially after convergence, compared to other models. Notably, all tasks provide only sparse rewards, CenRA addresses this challenge through the auxiliary dense rewards with meaningful information, effectively guiding learning. This mechanism not only distinguishes CenRA from other structurally shared methods, but also provides a targeted solution to the sparse-reward problem. Second, while baselines like PiCor and MCAL often show uneven progress across different tasks within the same domain, CenRA maintains well-balanced performance by showing relatively consistent learning progress and minimal variability across each four-task groups. This ensures that no single task dominates or lags behind, which is crucial in multi-task learning. Third, the CRA effectively enhances knowledge sharing among tasks. This is evident from the comparison with ReLara, which uses independent reward agents and lacks the mechanism for knowledge exchange. By extracting and distributing insights from one task to another, the CRA improves the learning efficiency of individual tasks, highlighting the advantages of integrated knowledge management.

\subsection{Knowledge Transfer to New Tasks}

\begin{figure*}[t]
\centering
\begin{subfigure}[b]{0.62\textwidth}
    \includegraphics[width=\textwidth]{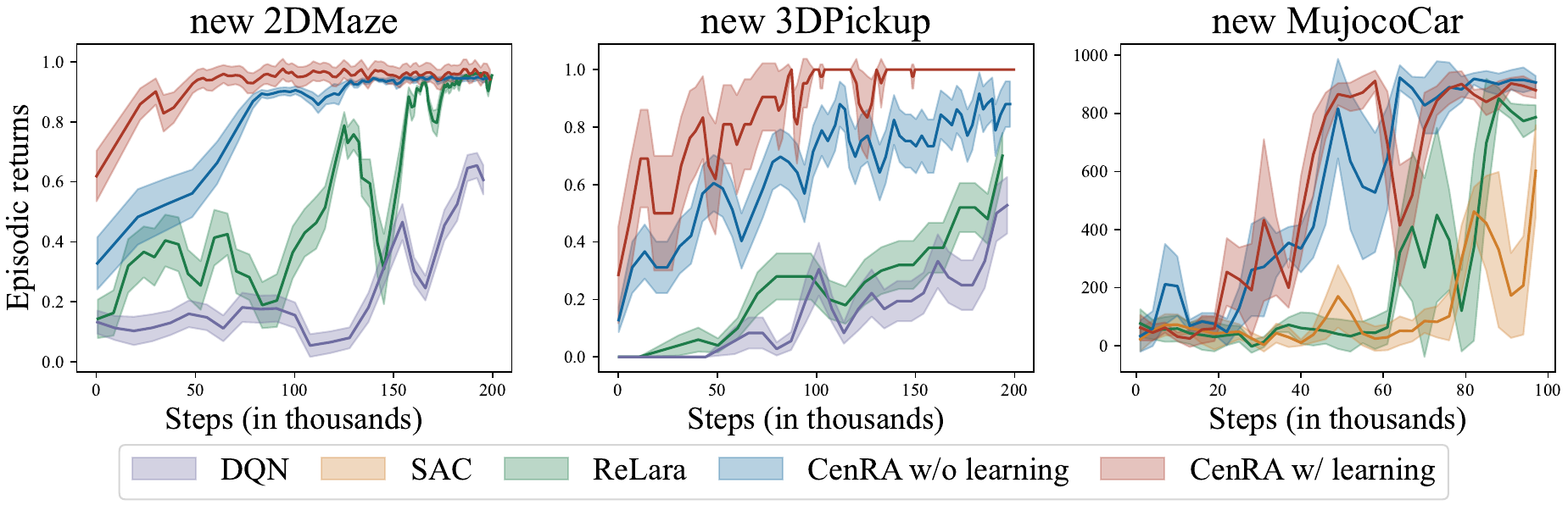}
    \caption{Comparison of the learning performance of CenRA with the baselines in new tasks in the \textit{2DMaze}, \textit{3DPickup} and \textit{MujocoCar} domains.}
    \label{fig:new-task}
\end{subfigure}
\hfill
\begin{subfigure}[b]{0.36\textwidth}
    \includegraphics[width=\textwidth]{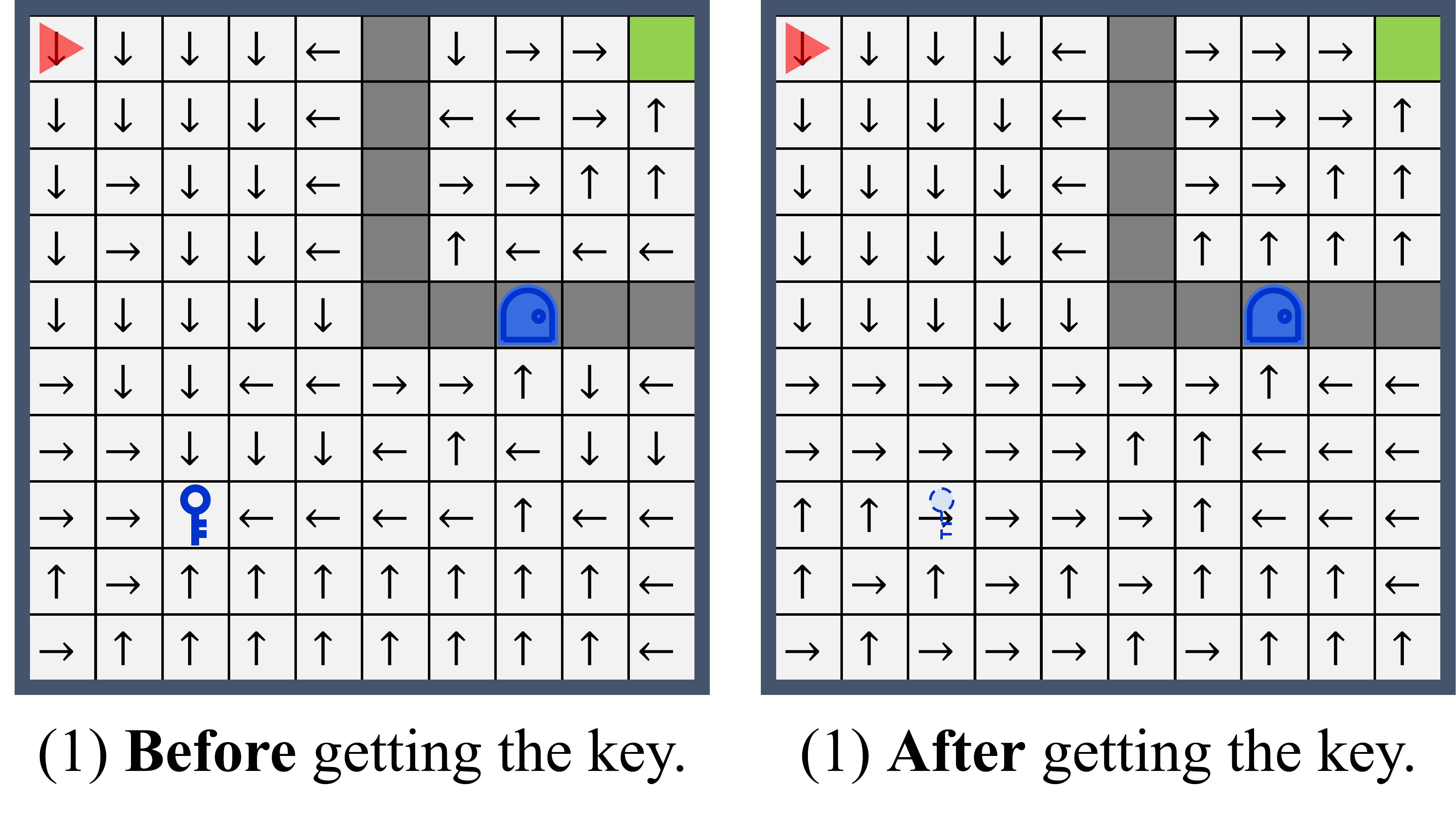}
    \caption{Actions yielding the highest knowledge rewards in a new \textit{2DMaze} task.}
    \label{fig:task-rewards}
\end{subfigure}
\caption{Experimental results for knowledge transfer to new tasks.}
\label{fig:new-task-results}
\end{figure*}

In this section, we assess the CRA's ability to transfer previously learned knowledge to unseen tasks. Specifically, we continue using the trained CRA model in Section~\ref{sec:comp}, while initializing new policy agents to tackle new tasks from the same domain. These include 5 test tasks for \textit{ML10-sparse} and \textit{ML50-sparse}, and 1 test task for each of the remaining domains, none of which were encountered during the initial training. For the CenRA, we explore two scenarios: (1) the CRA continues to be optimized in collaboration with the new policy agent (CenRA w/ learning); and (2) only the policy agent is updated while the CRA remains fixed, relying only on its previously acquired knowledge (CenRA w/o learning). We compare the two settings against the backbone algorithms and ReLara. In ReLara, the reward agent is trained anew without pre-learned knowledge. The results are presented in Figure~\ref{fig:new-task} and Table~\ref{tab:new-task}.

We observe that CenRA with further learning achieved rapid convergence, mainly due to the CRA's ability to retain previously acquired knowledge while continuing to adapt to new tasks through ongoing optimization. Remarkably, even without any additional training, CenRA still outperforms both ReLara, which requires training a new reward agent, and the backbone algorithms, which lack additional information. This advantage stems from the CRA's ability to encode and transfer environment-relevant knowledge, which can then be directly reused by new policy agents to guide their learning. Such knowledge transfer is particularly critical in our experiments involving challenging sparse-reward tasks. Without any external knowledge, learning would require extensive exploration. However, the CRA provides meaningful dense rewards that significantly accelerate the learning process, even during the initial phases.

\begin{table}[t]
\centering
\caption{Episodic returns (mean $\pm$ standard error) of all trained agents in the new tasks, tested over 100 episodes in each domain ($\uparrow$ higher is better).}
\label{tab:new-task}
\small
\setlength{\tabcolsep}{4pt}

\begin{tabular}{cccccc}
\toprule
Algorithm & \textit{ML10-sparse} & \textit{ML50-sparse} & \textit{2DMaze} & \textit{3DPickup} & \textit{MujocoCar} \\
\midrule
CenRA w/ learning  & \textbf{0.902 $\pm$ 0.021} & \textbf{0.824 $\pm$ 0.012} & \textbf{0.952 $\pm$ 0.010} & \textbf{0.963 $\pm$ 0.002} & \textbf{532.080 $\pm$ 1.610} \\
CenRA w/o learning & 0.887 $\pm$ 0.011 & 0.809 $\pm$ 0.009 & 0.894 $\pm$ 0.032 & 0.678 $\pm$ 0.003 & 524.727 $\pm$ 0.588 \\
ReLara             & 0.702 $\pm$ 0.086 & 0.612 $\pm$ 0.012 & 0.759 $\pm$ 0.056 & 0.263 $\pm$ 0.002 & 224.648 $\pm$ 0.492 \\
DQN/SAC            & 0.228 $\pm$ 0.105 & 0.210 $\pm$ 0.034 & 0.263 $\pm$ 0.084 & 0.158 $\pm$ 0.003 & 129.055 $\pm$ 0.296 \\
\bottomrule
\end{tabular}
\end{table}

To further demonstrate CenRA's transferability, we select the \textit{2DMaze} environment to visualize the knowledge provided by CRA when facing an unseen task. As shown in Figure~\ref{fig:task-rewards}, we plot the directions of actions that yield the maximum knowledge reward at each position, categorized into two scenarios: before and after obtaining the key. While some guidance in peripheral regions may appear slightly misaligned, most states receive reasonable rewards that align with human understanding. This demonstrates the effectiveness of knowledge transfer and with such dense rewards, the agent's adaptation to new tasks is able to be well supported. In addition, to better understand the knowledge learned by the CRA, we provide a case study in Appendix~\ref{app:case-study}, where we visualize the CRA-provided rewards. This further verifies that the CRA is capable of capturing domain-relevant, task-specific, and semantically meaningful signals across different tasks.

\begin{table}[t]
\centering
\caption{Comparison of CenRA with ablation of different batch sampling control weights.}
\label{tab:ablation}
\small
\setlength{\tabcolsep}{4pt}

\begin{tabular}{cccccc}
    \toprule
    \multirow{2}{*}{Algo.} & \multicolumn{4}{c}{\textit{2DMaze}} & \multirow{2}{*}{\shortstack{Var. $\downarrow$ \\ ($\times 10^{-2}$)}} \\
    \cmidrule(lr){2-5}
    & Maze \#1 & Maze \#2 & Maze \#3 & Maze \#4 & \\
    \midrule
    \textbf{CenRA} ($\alpha = 0.5$) & \textbf{0.893 $\pm$ 0.033} & 0.908 $\pm$ 0.022 & \textbf{0.924 $\pm$ 0.020} & \textbf{0.932 $\pm$ 0.020} & \underline{0.021} \\
    CenRA ($\alpha = 0.25$) & 0.889 $\pm$ 0.031 & 0.901 $\pm$ 0.025 & 0.915 $\pm$ 0.023 & 0.925 $\pm$ 0.024 & 0.065 \\
    CenRA ($\alpha = 0.75$) & 0.891 $\pm$ 0.030 & 0.905 $\pm$ 0.024 & 0.918 $\pm$ 0.021 & 0.928 $\pm$ 0.022 & 0.049 \\
    w/o $\bm{w}^{sim}$ ($\alpha = 0$) & 0.884 $\pm$ 0.033 & \textbf{0.922 $\pm$ 0.021} & 0.873 $\pm$ 0.041 & 0.820 $\pm$ 0.039 & 0.172 \\
    w/o $\bm{w}^{per}$ ($\alpha = 1$) & 0.758 $\pm$ 0.062 & 0.884 $\pm$ 0.030 & 0.824 $\pm$ 0.052 & 0.867 $\pm$ 0.020 & 0.284 \\
    w/o both & 0.632 $\pm$ 0.053 & 0.833 $\pm$ 0.041 & 0.629 $\pm$ 0.08 & 0.802 $\pm$ 0.054 & 1.235 \\
    \toprule
    \multirow{2}{*}{Algo.} & \multicolumn{4}{c}{\textit{3DPickup}} & \multirow{2}{*}{\shortstack{Var. $\downarrow$ \\ ($\times 10^{-2}$)}} \\
    \cmidrule(lr){2-5}
    & Ball & Cube & Key & Health kit & \\
    \midrule
    \textbf{CenRA} ($\alpha = 0.5$) & \textbf{0.951 $\pm$ 0.020} & 0.683 $\pm$ 0.090 & \textbf{0.795 $\pm$ 0.062} & 0.688 $\pm$ 0.067 & 1.570 \\
    CenRA ($\alpha = 0.25$) & 0.942 $\pm$ 0.023 & 0.671 $\pm$ 0.091 & 0.782 $\pm$ 0.066 & 0.715 $\pm$ 0.061 & 1.650 \\
    CenRA ($\alpha = 0.75$) & 0.938 $\pm$ 0.025 & 0.665 $\pm$ 0.095 & 0.775 $\pm$ 0.069 & 0.702 $\pm$ 0.065 & 1.723 \\
    w/o $\bm{w}^{sim}$ ($\alpha = 0$) & 0.822 $\pm$ 0.065 & \textbf{0.702 $\pm$ 0.093} & 0.704 $\pm$ 0.072 & \textbf{0.887 $\pm$ 0.038} & \underline{0.892} \\
    w/o $\bm{w}^{per}$ ($\alpha = 1$) & 0.779 $\pm$ 0.072 & 0.404 $\pm$ 0.093 & 0.631 $\pm$ 0.080 & 0.438 $\pm$ 0.102 & 3.051 \\
    w/o both & 0.811 $\pm$ 0.073 & 0.457 $\pm$ 0.058 & 0.483 $\pm$ 0.079 & 0.370 $\pm$ 0.079 & 3.796 \\
    \toprule
    \multirow{2}{*}{Algo.} & \multicolumn{4}{c}{\textit{MujocoCar}} & \multirow{2}{*}{\shortstack{Var. $\downarrow$ \\ ($\times 10^3$)}} \\
    \cmidrule(lr){2-5}
    & Target \#1 & Target \#2 & Target \#3 & Target \#4 & \\
    \midrule
    \textbf{CenRA} ($\alpha = 0.5$) & \textbf{588.221 $\pm$ 0.732} & \textbf{549.337 $\pm$ 0.640} & 447.743 $\pm$ 0.672 & \textbf{474.320 $\pm$ 0.657} & 4.244 \\
    CenRA ($\alpha = 0.25$) & 575.153 $\pm$ 0.740 & 538.912 $\pm$ 0.655 & 439.850 $\pm$ 0.680 & 462.116 $\pm$ 0.665 & 4.871 \\
    CenRA ($\alpha = 0.75$) & 580.431 $\pm$ 0.735 & 542.765 $\pm$ 0.648 & 441.033 $\pm$ 0.675 & 468.529 $\pm$ 0.660 & 4.533 \\
    w/o $\bm{w}^{sim}$ ($\alpha = 0$) & 319.926 $\pm$ 0.590 & 486.767 $\pm$ 0.712 & 332.506 $\pm$ 0.695 & 260.921 $\pm$ 0.544 & 9.288 \\
    w/o $\bm{w}^{per}$ ($\alpha = 1$) & 320.887 $\pm$ 0.891 & 355.325 $\pm$ 0.677 & 344.145 $\pm$ 0.872 & 308.215 $\pm$ 0.723 & \underline{0.457} \\
    w/o both & 57.532 $\pm$ 0.352 & 257.010 $\pm$ 0.677 & \textbf{677.285 $\pm$ 0.540} & 77.635 $\pm$ 0.255 & 82.720 \\
    \bottomrule
\end{tabular}
\end{table}

\subsection{Effect of Sampling Weight}

We conduct experiments to understand the effects of the information synchronization mechanism in the CenRA. Specifically, we compare the full CenRA model against five variants: (a) and (b) CenRA with different values of the balance factor $\alpha$, i.e., $\alpha = 0.25$ and $\alpha = 0.75$, to examine the impact of different weight combinations; (c) CenRA without the similarity weight $\bm{w}^{sim}$ (i.e., $\alpha = 0$); (d) CenRA without the performance weight $\bm{w}^{per}$ (i.e., $\alpha = 1$); and (e) CenRA without the entire sampling weight. To better illustrate the differences among tasks and highlight the role of sampling weights in task coordination and synchronization, we select the four-task domains, i.e., \textit{2DMaze}, \textit{3DPickup}, and \textit{MujocoCar}. The results are shown in Table~\ref{tab:ablation}.

The results indicate that the two weights, which control the allocation of samples drawn from each policy agent's experiences, mainly influence the overall learning performance. Specifically, the absence of sampling weight leads to unbalanced learning outcomes, which is observed by the increased variance in episodic returns across four tasks. In contrast, when both weights are incorporated, the learning process becomes notably more stable, indicating that the joint consideration of task similarity and learning progress is essential for coordinated optimization. While the full CenRA model does not always achieve the lowest variance, it consistently outperforms the other three ablation models regarding overall system performance.

Both weights play essential roles in information synchronization, with the performance weight $\bm{w}^{\mathrm{per}}$ having a more significant impact. It allows the CRA to focus more on policy agents that are underperforming or progressing slowly, ensuring balanced system-wide learning. Moreover, different choices of the balancing factor $\alpha$ emphasize distinct aspects of synchronization: a larger $\alpha$ highlights task similarity and promotes uniformity across related tasks, whereas a smaller $\alpha$ prioritizes compensating lagging tasks by amplifying the effect of $\bm{w}^{\mathrm{per}}$. This flexible weighting further enhances stability and adaptability, demonstrating that considering the overall learning performance of the multi-task system is a central objective that CenRA seeks to achieve.

\section{Discussion and Conclusion}
\label{sec:conclusion}

We propose a novel framework CenRA that integrates reward shaping into multi-task reinforcement learning. The framework shares domain knowledge across tasks to improve learning efficiency and effectively addresses the sparse-reward challenge. Specifically, the centralized reward agent (CRA) functions as a knowledge pool, responsible for distilling and distributing knowledge across tasks. Furthermore, the information synchronization mechanism mitigates imbalances in knowledge distribution, ensuring optimal system-wide performance. Experiments demonstrate that dense knowledge rewards generated by the CRA effectively guide policy learning, leading to faster convergence than baseline methods. CenRA also demonstrates superior and robust transferability to new tasks.

CenRA's main limitation is its requirement for consistent state and action dimensions across tasks. Future work could explore preprocessing techniques to adapt the framework to varying task structures, broadening its applicability. Additionally, the fixed trade-off between similarity weight and performance weight may not be ideal. A more flexible approach, such as adaptive weight regulation, could further enhance the framework. Moreover, the performance weight might favor underperforming tasks to achieve overall balance, but could limit the performance ceiling of high-performing tasks, indicating the need for a more effective trade-off mechanism.

\begin{ack}

This work was supported by an Academic Research Grant MOE-T1 251RES2408 and a Research Scholarship from the Ministry of Education, Singapore.

\end{ack}

\bibliography{reference}

\begin{thebibliography}{62}
\providecommand{\natexlab}[1]{#1}
\providecommand{\url}[1]{\texttt{#1}}
\expandafter\ifx\csname urlstyle\endcsname\relax
  \providecommand{\doi}[1]{doi: #1}\else
  \providecommand{\doi}{doi: \begingroup \urlstyle{rm}\Url}\fi

\bibitem[Ammar et~al.(2014)Ammar, Eaton, Ruvolo, and Taylor]{ammar2014online}
Haitham~Bou Ammar, Eric Eaton, Paul Ruvolo, and Matthew Taylor.
\newblock Online multi-task learning for policy gradient methods.
\newblock In \emph{International conference on machine learning}, pages 1206--1214. PMLR, 2014.

\bibitem[Aradi(2020)]{aradi2020survey}
Szil{\'a}rd Aradi.
\newblock Survey of deep reinforcement learning for motion planning of autonomous vehicles.
\newblock \emph{IEEE Transactions on Intelligent Transportation Systems}, 23\penalty0 (2):\penalty0 740--759, 2020.

\bibitem[Bai et~al.(2023)Bai, Zhang, Tao, Wu, Wang, and Xu]{picor:bai2023picor}
Fengshuo Bai, Hongming Zhang, Tianyang Tao, Zhiheng Wu, Yanna Wang, and Bo~Xu.
\newblock Picor: Multi-task deep reinforcement learning with policy correction.
\newblock In \emph{Proceedings of the AAAI Conference on Artificial Intelligence}, volume~37, pages 6728--6736, 2023.

\bibitem[Bellemare et~al.(2016)Bellemare, Srinivasan, Ostrovski, Schaul, Saxton, and Munos]{rs-explo:bellemare2016unifying}
Marc Bellemare, Sriram Srinivasan, Georg Ostrovski, Tom Schaul, David Saxton, and Remi Munos.
\newblock Unifying count-based exploration and intrinsic motivation.
\newblock \emph{Advances in Neural Information Processing Systems}, 29, 2016.

\bibitem[Burda et~al.(2018)Burda, Edwards, Storkey, and Klimov]{rs-novelty:burda2018exploration}
Yuri Burda, Harrison Edwards, Amos Storkey, and Oleg Klimov.
\newblock Exploration by random network distillation.
\newblock In \emph{International Conference on Learning Representations}, 2018.

\bibitem[Caruana(1993)]{mtrl:caruana1993multitask}
R~Caruana.
\newblock Multitask learning: A knowledge-based source of inductive bias1.
\newblock In \emph{Proceedings of the Tenth International Conference on Machine Learning}, pages 41--48. Citeseer, 1993.

\bibitem[Chen et~al.(2018)Chen, Badrinarayanan, Lee, and Rabinovich]{chen2018gradnorm}
Zhao Chen, Vijay Badrinarayanan, Chen-Yu Lee, and Andrew Rabinovich.
\newblock Gradnorm: Gradient normalization for adaptive loss balancing in deep multitask networks.
\newblock In \emph{International conference on machine learning}, pages 794--803. PMLR, 2018.

\bibitem[Cheng et~al.(2023)Cheng, Dong, Cai, and Sun]{cheng2023multi}
Guangran Cheng, Lu~Dong, Wenzhe Cai, and Changyin Sun.
\newblock Multi-task reinforcement learning with attention-based mixture of experts.
\newblock \emph{IEEE Robotics and Automation Letters}, 8\penalty0 (6):\penalty0 3812--3819, 2023.

\bibitem[Chevalier-Boisvert et~al.(2024)Chevalier-Boisvert, Dai, Towers, Perez-Vicente, Willems, Lahlou, Pal, Castro, and Terry]{env-mini:chevalier2024minigrid}
Maxime Chevalier-Boisvert, Bolun Dai, Mark Towers, Rodrigo Perez-Vicente, Lucas Willems, Salem Lahlou, Suman Pal, Pablo~Samuel Castro, and Jordan Terry.
\newblock Minigrid \& miniworld: Modular \& customizable reinforcement learning environments for goal-oriented tasks.
\newblock \emph{Advances in Neural Information Processing Systems}, 36, 2024.

\bibitem[D'Eramo et~al.(2020)D'Eramo, Tateo, Bonarini, Restelli, and Peters]{dsharing}
Carlo D'Eramo, Davide Tateo, Andrea Bonarini, Marcello Restelli, and Jan Peters.
\newblock Sharing knowledge in multi-task deep reinforcement learning.
\newblock In \emph{International Conference on Learning Representations}, 2020.

\bibitem[Devidze et~al.(2022)Devidze, Kamalaruban, and Singla]{rs-explo:devidze2022exploration}
Rati Devidze, Parameswaran Kamalaruban, and Adish Singla.
\newblock Exploration-guided reward shaping for reinforcement learning under sparse rewards.
\newblock \emph{Advances in Neural Information Processing Systems}, 35:\penalty0 5829--5842, 2022.

\bibitem[Devin et~al.(2017)Devin, Gupta, Darrell, Abbeel, and Levine]{devin2017learning}
Coline Devin, Abhishek Gupta, Trevor Darrell, Pieter Abbeel, and Sergey Levine.
\newblock Learning modular neural network policies for multi-task and multi-robot transfer.
\newblock In \emph{2017 IEEE international conference on robotics and automation (ICRA)}, pages 2169--2176. IEEE, 2017.

\bibitem[Finn et~al.(2017)Finn, Abbeel, and Levine]{finn2017model}
Chelsea Finn, Pieter Abbeel, and Sergey Levine.
\newblock Model-agnostic meta-learning for fast adaptation of deep networks.
\newblock In \emph{International conference on machine learning}, pages 1126--1135. PMLR, 2017.

\bibitem[Fujimoto et~al.(2018)Fujimoto, Hoof, and Meger]{fujimoto2018addressing}
Scott Fujimoto, Herke Hoof, and David Meger.
\newblock Addressing function approximation error in actor-critic methods.
\newblock In \emph{International conference on machine learning}, pages 1587--1596. PMLR, 2018.

\bibitem[Haarnoja et~al.(2018{\natexlab{a}})Haarnoja, Zhou, Abbeel, and Levine]{sac:haarnoja2018soft}
Tuomas Haarnoja, Aurick Zhou, Pieter Abbeel, and Sergey Levine.
\newblock Soft actor-critic: Off-policy maximum entropy deep reinforcement learning with a stochastic actor.
\newblock In \emph{International Conference on Machine Learning}, pages 1861--1870. PMLR, 2018{\natexlab{a}}.

\bibitem[Haarnoja et~al.(2018{\natexlab{b}})Haarnoja, Zhou, Hartikainen, Tucker, Ha, Tan, Kumar, Zhu, Gupta, Abbeel, et~al.]{sac-app:haarnoja2018soft}
Tuomas Haarnoja, Aurick Zhou, Kristian Hartikainen, George Tucker, Sehoon Ha, Jie Tan, Vikash Kumar, Henry Zhu, Abhishek Gupta, Pieter Abbeel, et~al.
\newblock Soft actor-critic algorithms and applications.
\newblock \emph{arXiv preprint arXiv:1812.05905}, 2018{\natexlab{b}}.

\bibitem[Hansen et~al.(2024)Hansen, Su, and Wang]{TD-MPC2:hansen2024td}
Nicklas Hansen, Hao Su, and Xiaolong Wang.
\newblock Td-mpc2: Scalable, robust world models for continuous control.
\newblock In \emph{The Twelfth International Conference on Learning Representations}, 2024.

\bibitem[He et~al.(2024)He, Li, Zang, Fu, Fu, Xing, and Cheng]{he2024not}
Jinmin He, Kai Li, Yifan Zang, Haobo Fu, Qiang Fu, Junliang Xing, and Jian Cheng.
\newblock Not all tasks are equally difficult: Multi-task deep reinforcement learning with dynamic depth routing.
\newblock In \emph{Proceedings of the AAAI Conference on Artificial Intelligence}, volume~38, pages 12376--12384, 2024.

\bibitem[Hessel et~al.(2019)Hessel, Soyer, Espeholt, Czarnecki, Schmitt, and Van~Hasselt]{hessel2019multi}
Matteo Hessel, Hubert Soyer, Lasse Espeholt, Wojciech Czarnecki, Simon Schmitt, and Hado Van~Hasselt.
\newblock Multi-task deep reinforcement learning with popart.
\newblock In \emph{Proceedings of the AAAI Conference on Artificial Intelligence}, volume~33, pages 3796--3803, 2019.

\bibitem[Hong et~al.(2021)Hong, Yoon, and Kim]{hong2021structure}
Sunghoon Hong, Deunsol Yoon, and Kee-Eung Kim.
\newblock Structure-aware transformer policy for inhomogeneous multi-task reinforcement learning.
\newblock In \emph{International Conference on Learning Representations}, 2021.

\bibitem[Huang et~al.(2022)Huang, Dossa, Ye, Braga, Chakraborty, Mehta, and Ara{\~A}{\v{s}}jo]{cleanrl:huang2022cleanrl}
Shengyi Huang, Rousslan Fernand~Julien Dossa, Chang Ye, Jeff Braga, Dipam Chakraborty, Kinal Mehta, and Jo{\~A}{\c{G}}o~GM Ara{\~A}{\v{s}}jo.
\newblock Cleanrl: High-quality single-file implementations of deep reinforcement learning algorithms.
\newblock \emph{Journal of Machine Learning Research}, 23\penalty0 (274):\penalty0 1--18, 2022.

\bibitem[Ji et~al.(2023)Ji, Zhang, Zhou, Pan, Huang, Sun, Geng, Zhong, Dai, and Yang]{env-car:ji2023safety}
Jiaming Ji, Borong Zhang, Jiayi Zhou, Xuehai Pan, Weidong Huang, Ruiyang Sun, Yiran Geng, Yifan Zhong, Josef Dai, and Yaodong Yang.
\newblock Safety gymnasium: A unified safe reinforcement learning benchmark.
\newblock \emph{Advances in Neural Information Processing Systems}, 36, 2023.

\bibitem[Kober et~al.(2013)Kober, Bagnell, and Peters]{kober2013reinforcement}
Jens Kober, J~Andrew Bagnell, and Jan Peters.
\newblock Reinforcement learning in robotics: A survey.
\newblock \emph{The International Journal of Robotics Research}, 32\penalty0 (11):\penalty0 1238--1274, 2013.

\bibitem[Konda and Tsitsiklis(1999)]{konda1999actor}
Vijay Konda and John Tsitsiklis.
\newblock Actor-critic algorithms.
\newblock \emph{Advances in neural information processing systems}, 12, 1999.

\bibitem[Ladosz et~al.(2022)Ladosz, Weng, Kim, and Oh]{expl-survey:ladosz2022exploration}
Pawel Ladosz, Lilian Weng, Minwoo Kim, and Hyondong Oh.
\newblock Exploration in deep reinforcement learning: A survey.
\newblock \emph{Information Fusion}, 85:\penalty0 1--22, 2022.

\bibitem[Lample and Chaplot(2017)]{lample2017playing}
Guillaume Lample and Devendra~Singh Chaplot.
\newblock Playing fps games with deep reinforcement learning.
\newblock In \emph{Proceedings of the AAAI conference on artificial intelligence}, volume~31, 2017.

\bibitem[Lan et~al.(2023)Lan, Zhang, Yi, Guo, Peng, Gao, Wu, Chen, Du, Hu, et~al.]{lan2024contrastive}
Siming Lan, Rui Zhang, Qi~Yi, Jiaming Guo, Shaohui Peng, Yunkai Gao, Fan Wu, Ruizhi Chen, Zidong Du, Xing Hu, et~al.
\newblock Contrastive modules with temporal attention for multi-task reinforcement learning.
\newblock \emph{Advances in Neural Information Processing Systems}, 36, 2023.

\bibitem[Luo et~al.(2024)Luo, Ma, Shi, and Gan]{luo4837239gfanc}
Zhengding Luo, Haozhe Ma, Dongyuan Shi, and Woon-Seng Gan.
\newblock Gfanc-rl: Reinforcement learning-based generative fixed-filter active noise control.
\newblock \emph{Available at SSRN 4837239}, 2024.

\bibitem[Ma et~al.(2023)Ma, Vo, and Leong]{mine:ma2023hierarchical}
Haozhe Ma, Thanh~Vinh Vo, and Tze-Yun Leong.
\newblock Hierarchical reinforcement learning with human-ai collaborative sub-goals optimization.
\newblock In \emph{Proceedings of the 2023 international conference on autonomous agents and multiagent systems}, pages 2310--2312, 2023.

\bibitem[Ma et~al.(2024{\natexlab{a}})Ma, Sima, Vo, Fu, and Leong]{relara:ma2024reward}
Haozhe Ma, Kuankuan Sima, Thanh~Vinh Vo, Di~Fu, and Tze-Yun Leong.
\newblock Reward shaping for reinforcement learning with an assistant reward agent.
\newblock In \emph{Forty-first International Conference on Machine Learning}. PMLR, 2024{\natexlab{a}}.

\bibitem[Ma et~al.(2024{\natexlab{b}})Ma, Vo, and Leong]{ma2024mixed}
Haozhe Ma, Thanh~Vinh Vo, and Tze-Yun Leong.
\newblock Mixed-initiative bayesian sub-goal optimization in hierarchical reinforcement learning.
\newblock In \emph{Proceedings of the 23rd International Conference on Autonomous Agents and Multiagent Systems}, pages 1328--1336, 2024{\natexlab{b}}.

\bibitem[Ma et~al.(2025{\natexlab{a}})Ma, Li, Lim, Luo, Vo, and Leong]{mine:durnd}
Haozhe Ma, Fangling Li, Jing~Yu Lim, Zhengding Luo, Thanh~Vinh Vo, and Tze-Yun Leong.
\newblock Catching two birds with one stone: Reward shaping with dual random networks for balancing exploration and exploitation.
\newblock In \emph{Forty-second International Conference on Machine Learning}, 2025{\natexlab{a}}.

\bibitem[Ma et~al.(2025{\natexlab{b}})Ma, Luo, Vo, Sima, and Leong]{ma2024highly}
Haozhe Ma, Zhengding Luo, Thanh~Vinh Vo, Kuankuan Sima, and Tze-Yun Leong.
\newblock Highly efficient self-adaptive reward shaping for reinforcement learning.
\newblock In \emph{Thirteenth International Conference on Learning Representations}, 2025{\natexlab{b}}.

\bibitem[Mavor-Parker et~al.(2022)Mavor-Parker, Young, Barry, and Griffin]{rs-curi:mavor2022stay}
Augustine Mavor-Parker, Kimberly Young, Caswell Barry, and Lewis Griffin.
\newblock How to stay curious while avoiding noisy tvs using aleatoric uncertainty estimation.
\newblock In \emph{International Conference on Machine Learning}, pages 15220--15240. PMLR, 2022.

\bibitem[Memarian et~al.(2021)Memarian, Goo, Lioutikov, Niekum, and Topcu]{memarian2021self}
Farzan Memarian, Wonjoon Goo, Rudolf Lioutikov, Scott Niekum, and Ufuk Topcu.
\newblock Self-supervised online reward shaping in sparse-reward environments.
\newblock In \emph{2021 IEEE/RSJ International Conference on Intelligent Robots and Systems (IROS)}, pages 2369--2375. IEEE, 2021.

\bibitem[Mguni et~al.(2023)Mguni, Jafferjee, Wang, Perez-Nieves, Song, Tong, Taylor, Yang, Dai, Chen, et~al.]{rs-multi:mguni2023learning}
David Mguni, Taher Jafferjee, Jianhong Wang, Nicolas Perez-Nieves, Wenbin Song, Feifei Tong, Matthew Taylor, Tianpei Yang, Zipeng Dai, Hui Chen, et~al.
\newblock Learning to shape rewards using a game of two partners.
\newblock In \emph{AAAI Conference on Artificial Intelligence}, volume~37, pages 11604--11612, 2023.

\bibitem[Mnih et~al.(2015)Mnih, Kavukcuoglu, Silver, Rusu, Veness, Bellemare, Graves, Riedmiller, Fidjeland, Ostrovski, et~al.]{dqn:mnih2015human}
Volodymyr Mnih, Koray Kavukcuoglu, David Silver, Andrei~A Rusu, Joel Veness, Marc~G Bellemare, Alex Graves, Martin Riedmiller, Andreas~K Fidjeland, Georg Ostrovski, et~al.
\newblock Human-level control through deep reinforcement learning.
\newblock \emph{Nature}, 518\penalty0 (7540):\penalty0 529--533, 2015.

\bibitem[Mysore et~al.(2022)Mysore, Cheng, Zhao, Saenko, and Wu]{mcal:mysore2022multi}
Siddharth Mysore, George Cheng, Yunqi Zhao, Kate Saenko, and Meng Wu.
\newblock Multi-critic actor learning: Teaching rl policies to act with style.
\newblock In \emph{International Conference on Learning Representations}, 2022.

\bibitem[Ostrovski et~al.(2017)Ostrovski, Bellemare, Oord, and Munos]{rs-explo:ostrovski2017count}
Georg Ostrovski, Marc~G Bellemare, A{\"a}ron Oord, and R{\'e}mi Munos.
\newblock Count-based exploration with neural density models.
\newblock In \emph{International Conference on Machine Learning}, pages 2721--2730. PMLR, 2017.

\bibitem[Ouyang et~al.(2022)Ouyang, Wu, Jiang, Almeida, Wainwright, Mishkin, Zhang, Agarwal, Slama, Ray, et~al.]{ouyang2022training}
Long Ouyang, Jeffrey Wu, Xu~Jiang, Diogo Almeida, Carroll Wainwright, Pamela Mishkin, Chong Zhang, Sandhini Agarwal, Katarina Slama, Alex Ray, et~al.
\newblock Training language models to follow instructions with human feedback.
\newblock \emph{Advances in neural information processing systems}, 35:\penalty0 27730--27744, 2022.

\bibitem[Parisotto et~al.(2016)Parisotto, Ba, and Salakhutdinov]{parisotto16_actormimic}
Emilio Parisotto, Jimmy Ba, and Ruslan Salakhutdinov.
\newblock Actor-mimic: Deep multitask and transfer reinforcement learning.
\newblock In \emph{International Conference on Learning Representations}, 2016.

\bibitem[Pathak et~al.(2017)Pathak, Agrawal, Efros, and Darrell]{rs-curi:pathak2017curiosity}
Deepak Pathak, Pulkit Agrawal, Alexei~A Efros, and Trevor Darrell.
\newblock Curiosity-driven exploration by self-supervised prediction.
\newblock In \emph{International Conference on Machine Learning}, pages 2778--2787. PMLR, 2017.

\bibitem[Rusu et~al.(2016)Rusu, Colmenarejo, Gulcehre, Desjardins, Kirkpatrick, Pascanu, Mnih, Kavukcuoglu, and Hadsell]{rusu2015policy}
Andrei~A Rusu, Sergio~Gomez Colmenarejo, Caglar Gulcehre, Guillaume Desjardins, James Kirkpatrick, Razvan Pascanu, Volodymyr Mnih, Koray Kavukcuoglu, and Raia Hadsell.
\newblock Policy distillation.
\newblock In \emph{International Conference on Learning Representations}, 2016.

\bibitem[Shinn et~al.(2023)Shinn, Cassano, Gopinath, Narasimhan, and Yao]{shinn2024reflexion}
Noah Shinn, Federico Cassano, Ashwin Gopinath, Karthik Narasimhan, and Shunyu Yao.
\newblock Reflexion: Language agents with verbal reinforcement learning.
\newblock \emph{Advances in Neural Information Processing Systems}, 36:\penalty0 8634--8652, 2023.

\bibitem[Sodhani et~al.(2021)Sodhani, Zhang, and Pineau]{sodhani2021multi}
Shagun Sodhani, Amy Zhang, and Joelle Pineau.
\newblock Multi-task reinforcement learning with context-based representations.
\newblock In \emph{International Conference on Machine Learning}, pages 9767--9779. PMLR, 2021.

\bibitem[Sorg et~al.(2010{\natexlab{a}})Sorg, Lewis, and Singh]{reward-idea:sorg2010reward}
Jonathan Sorg, Richard~L Lewis, and Satinder Singh.
\newblock Reward design via online gradient ascent.
\newblock \emph{Advances in Neural Information Processing Systems}, 23, 2010{\natexlab{a}}.

\bibitem[Sorg et~al.(2010{\natexlab{b}})Sorg, Singh, and Lewis]{reward-idea:sorg2010internal}
Jonathan Sorg, Satinder~P Singh, and Richard~L Lewis.
\newblock Internal rewards mitigate agent boundedness.
\newblock In \emph{International Conference on Machine Learning}, pages 1007--1014, 2010{\natexlab{b}}.

\bibitem[Sun et~al.(2022)Sun, Zhang, Xu, and Tomizuka]{sun2022paco}
Lingfeng Sun, Haichao Zhang, Wei Xu, and Masayoshi Tomizuka.
\newblock Paco: Parameter-compositional multi-task reinforcement learning.
\newblock \emph{Advances in Neural Information Processing Systems}, 35:\penalty0 21495--21507, 2022.

\bibitem[Tang et~al.(2017)Tang, Houthooft, Foote, Stooke, Xi~Chen, Duan, Schulman, DeTurck, and Abbeel]{rs-explo:tang2017exploration}
Haoran Tang, Rein Houthooft, Davis Foote, Adam Stooke, OpenAI Xi~Chen, Yan Duan, John Schulman, Filip DeTurck, and Pieter Abbeel.
\newblock \# exploration: A study of count-based exploration for deep reinforcement learning.
\newblock \emph{Advances in Neural Information Processing Systems}, 30, 2017.

\bibitem[Teh et~al.(2017)Teh, Bapst, Czarnecki, Quan, Kirkpatrick, Hadsell, Heess, and Pascanu]{distral:teh2017distral}
Yee Teh, Victor Bapst, Wojciech~M Czarnecki, John Quan, James Kirkpatrick, Raia Hadsell, Nicolas Heess, and Razvan Pascanu.
\newblock Distral: Robust multitask reinforcement learning.
\newblock \emph{Advances in neural information processing systems}, 30, 2017.

\bibitem[Vaswani et~al.(2017)Vaswani, Shazeer, Parmar, Uszkoreit, Jones, Gomez, Kaiser, and Polosukhin]{vaswani2017attention}
Ashish Vaswani, Noam Shazeer, Niki Parmar, Jakob Uszkoreit, Llion Jones, Aidan~N Gomez, {\L}ukasz Kaiser, and Illia Polosukhin.
\newblock Attention is all you need.
\newblock \emph{Advances in neural information processing systems}, 30, 2017.

\bibitem[Vuong et~al.(2019)Vuong, Nguyen, Nguyen, Bui, Kieu, Ta, Tran, and Le]{vuong2019sharing}
Tung-Long Vuong, Do-Van Nguyen, Tai-Long Nguyen, Cong-Minh Bui, Hai-Dang Kieu, Viet-Cuong Ta, Quoc-Long Tran, and Thanh-Ha Le.
\newblock Sharing experience in multitask reinforcement learning.
\newblock In \emph{International Joint Conference on Artificial Intelligence}, pages 3642--3648, 2019.

\bibitem[Wan et~al.(2020)Wan, Gangwani, and Peng]{wan2020mutual}
Michael Wan, Tanmay Gangwani, and Jian Peng.
\newblock Mutual information based knowledge transfer under state-action dimension mismatch.
\newblock In \emph{Conference on Uncertainty in Artificial Intelligence}, pages 1218--1227. PMLR, 2020.

\bibitem[Xu et~al.(2024)Xu, Li, and Ren]{pmlr-v235-xu24o}
Tengye Xu, Zihao Li, and Qinyuan Ren.
\newblock Meta-reinforcement learning robust to distributional shift via performing lifelong in-context learning.
\newblock In \emph{Proceedings of the 41st International Conference on Machine Learning}, volume 235 of \emph{Proceedings of Machine Learning Research}, pages 55112--55125. PMLR, 2024.

\bibitem[Xu et~al.(2020)Xu, Wu, Che, Tang, and Ye]{xu2020knowledge}
Zhiyuan Xu, Kun Wu, Zhengping Che, Jian Tang, and Jieping Ye.
\newblock Knowledge transfer in multi-task deep reinforcement learning for continuous control.
\newblock \emph{Advances in Neural Information Processing Systems}, 33:\penalty0 15146--15155, 2020.

\bibitem[Yang et~al.(2020)Yang, Xu, Wu, and Wang]{yang2020multi}
Ruihan Yang, Huazhe Xu, Yi~Wu, and Xiaolong Wang.
\newblock Multi-task reinforcement learning with soft modularization.
\newblock \emph{Advances in Neural Information Processing Systems}, 33:\penalty0 4767--4777, 2020.

\bibitem[Yang et~al.(2017)Yang, Merrick, Abbass, and Jin]{yang2017multi}
Zhaoyang Yang, Kathryn~E Merrick, Hussein~A Abbass, and Lianwen Jin.
\newblock Multi-task deep reinforcement learning for continuous action control.
\newblock In \emph{International Joint Conference on Artificial Intelligence}, volume~17, pages 3301--3307, 2017.

\bibitem[Yin and Pan(2017)]{yin2017knowledge}
Haiyan Yin and Sinno Pan.
\newblock Knowledge transfer for deep reinforcement learning with hierarchical experience replay.
\newblock In \emph{Proceedings of the AAAI Conference on Artificial Intelligence}, volume~31, 2017.

\bibitem[Yu et~al.(2020{\natexlab{a}})Yu, Kumar, Gupta, Levine, Hausman, and Finn]{yu2020gradient}
Tianhe Yu, Saurabh Kumar, Abhishek Gupta, Sergey Levine, Karol Hausman, and Chelsea Finn.
\newblock Gradient surgery for multi-task learning.
\newblock \emph{Advances in Neural Information Processing Systems}, 33:\penalty0 5824--5836, 2020{\natexlab{a}}.

\bibitem[Yu et~al.(2020{\natexlab{b}})Yu, Quillen, He, Julian, Hausman, Finn, and Levine]{env-mw:yu2020meta}
Tianhe Yu, Deirdre Quillen, Zhanpeng He, Ryan Julian, Karol Hausman, Chelsea Finn, and Sergey Levine.
\newblock Meta-world: A benchmark and evaluation for multi-task and meta reinforcement learning.
\newblock In \emph{Conference on robot learning}, pages 1094--1100. PMLR, 2020{\natexlab{b}}.

\bibitem[Zeng et~al.(2021)Zeng, Anwar, Doan, Raychowdhury, and Romberg]{zeng2021decentralized}
Sihan Zeng, Malik~Aqeel Anwar, Thinh~T Doan, Arijit Raychowdhury, and Justin Romberg.
\newblock A decentralized policy gradient approach to multi-task reinforcement learning.
\newblock In \emph{Uncertainty in Artificial Intelligence}, pages 1002--1012. PMLR, 2021.

\bibitem[Zhang et~al.(2021)Zhang, Feng, and Hou]{sc:zhang2021multi}
Gengzhi Zhang, Liang Feng, and Yaqing Hou.
\newblock Multi-task actor-critic with knowledge transfer via a shared critic.
\newblock In \emph{Asian Conference on Machine Learning}, pages 580--593. PMLR, 2021.

\end{thebibliography}

\newpage
\appendix

\section{Mutli-Task Experimental Configurations}
\label{app:tasks}

We conduct experiments in four domains with multiple tasks: \textit{Meta-World}, \textit{2DMaze}, \textit{3DPickup}, and \textit{MujocoCar}. The detailed configurations of each task are illustrated in Figure~\ref{fig:appendix-tasks}. The \textit{Meta-World} tasks illustration is adapted from~\citep{env-mw:yu2020meta}.

\begin{figure}[h]
    \centering
    \includegraphics[width=0.94\linewidth]{Images/appendix/task-configs.pdf}
    \caption{Illustration of multiple tasks in different domains in our experiments.}
    \label{fig:appendix-tasks}
\end{figure}

\section{Network Structures and Hyperparameters}
\label{app:configs}

\subsection{Network Structures}

\begin{figure}[h!]
    \centering
    \includegraphics[width=\linewidth]{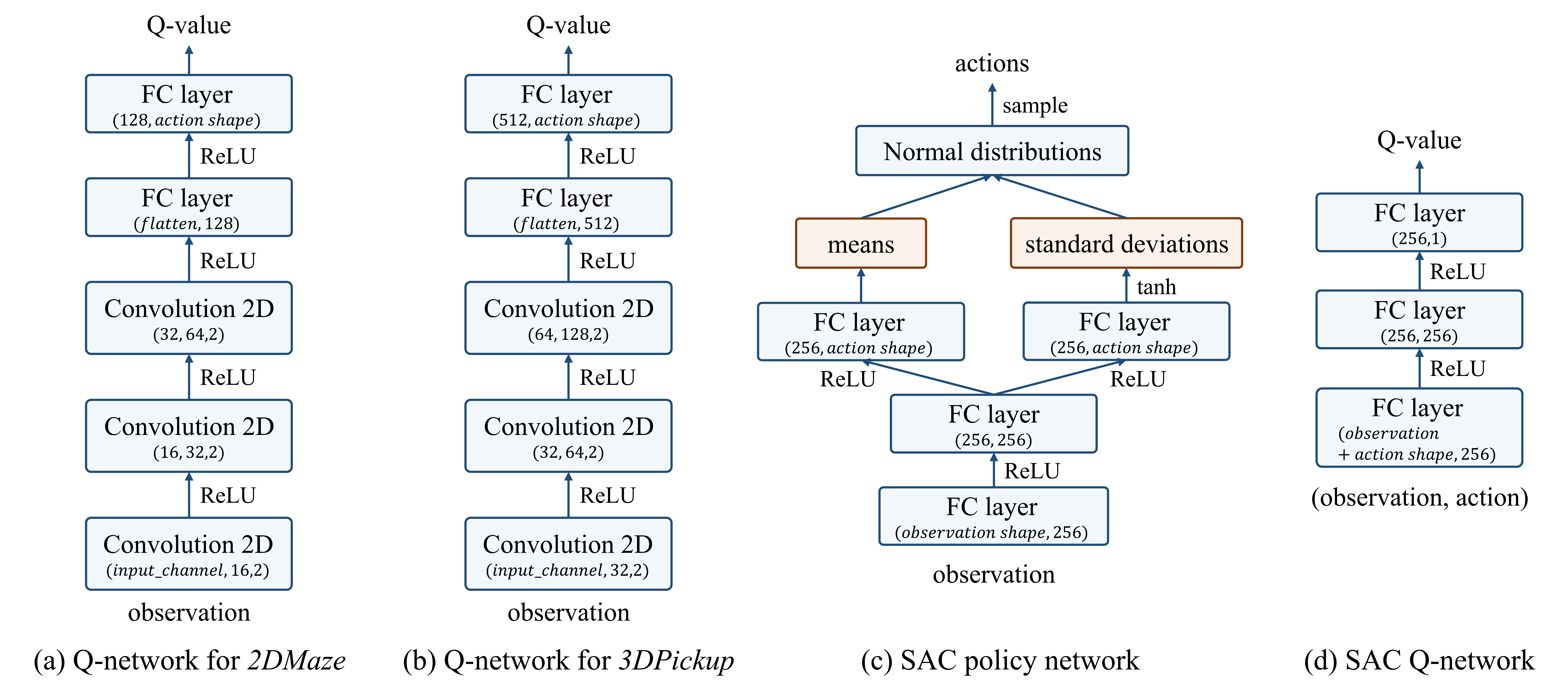}
    \caption{The structures of neural networks in our implementation.}
    \label{fig:appendix-networks}
\end{figure}

Figure~\ref{fig:appendix-networks} illustrates the structures of all networks employed in our experiments.

\subsection{Hyperparameters}

We have observed that CenRA demonstrated high robustness and was not sensitive to hyperparameter choices. Table~\ref{tab:appendix-hyperparameters} shows the hyperparameters we used in all the experiments.

\begin{table*}[h]
\centering        
\small
\caption{The hyperparameters of CenRA used in our experiments.}
\label{tab:appendix-hyperparameters}

\setlength{\tabcolsep}{10pt}
\begin{tabular}{ccc}
    \toprule
    Module & Hyperparameters & Values \\
    \midrule
    \multirow{8}{*}{\shortstack{Centralized Reward Agent \\ $\mathcal{A}^{rwd}$}} & discounted factor $\gamma$ & 0.99 \\
    & batch size & 256 \\
    & actor module learning rate & $3 \times 10^{-4}$ \\
    & critic module learning rate & $1 \times 10^{-3}$ \\
    & policy networks update frequency (steps) & $2$ \\
    & target networks update frequency (steps) & $1$ \\
    & target networks soft update weight $\tau$ & $5 \times 10^{-3}$ \\
    & burn-in steps & $5000$ \\
    \midrule
    \multirow{5}{*}{\shortstack{Policy Agent $\mathcal{A}^{pol}_i$ \\ (DQN Agent)}} & knowledge reward weight $\lambda$ & 0.5 \\
    & discounted factor $\gamma$ & 0.99 \\
    & replay buffer size $|\mathcal{D}_i|$ & $1 \times 10^6$ \\ 
    & batch size & 128 \\
    & burn-in steps & $10000$ \\
    \midrule
    \multirow{11}{*}{\shortstack{Policy Agent $\mathcal{A}^{pol}_i$ \\ (SAC Agent)}} & knowledge reward weight $\lambda$ & 0.5 \\
    & discounted factor $\gamma$ & 0.99 \\
    & replay buffer size $|\mathcal{D}_i|$ & $1 \times 10^6$ \\ 
    & batch size & 256 \\
    & actor module learning rate & $3 \times 10^{-4}$ \\
    & critic module learning rate & $1 \times 10^{-3}$ \\
    & SAC entropy term factor $\alpha$ learning rate & $1 \times 10^{-4}$ \\
    & policy networks update frequency (steps) & $2$ \\
    & target networks update frequency (steps) & $1$ \\
    & target networks soft update weight $\tau$ & $5 \times 10^{-3}$ \\
    & burn-in steps & $10000$ \\
    \bottomrule
\end{tabular}
\end{table*}

\subsection{Computing Resources}

The experiments in this paper were conducted on a computing cluster, with the detailed hardware configurations listed in Table~\ref{tab:appendix-resources}.

\begin{table*}[h!]
\centering
\small
\caption{The computing resources used in the experiments.}
\label{tab:appendix-resources}

\begin{tabular}{cc}
    \toprule
    Component & Specification \\
    \midrule
    Operating System (OS) & Ubuntu 20.04 \\
    Central Processing Unit (CPU) & 2x Intel Xeon Gold 6326 \\
    Random Access Memory (RAM) & 256GB \\
    Graphics Processing Unit (GPU) & 1x NVIDIA A100 20GB \\
    Brand & Supermicro 2022 \\
    \bottomrule
\end{tabular}
\vspace{-18pt}
\end{table*}

\section{What Has the Centralized Reward Agent Learned?}
\label{app:case-study}

\begin{figure}[h!]
\centering
\begin{subfigure}[b]{0.48\textwidth}
    \centering
    \includegraphics[width=0.9\textwidth]{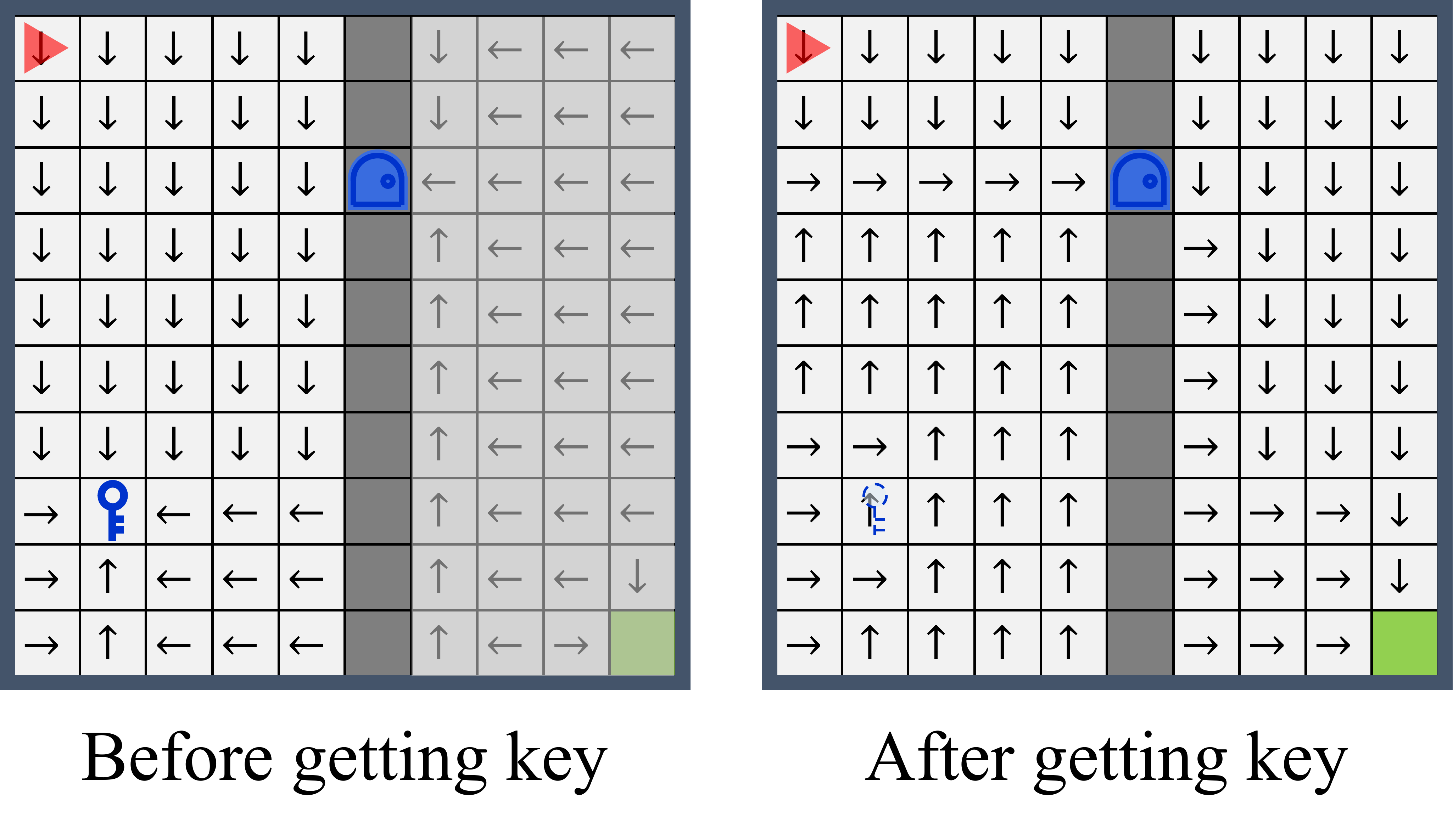}
    \vspace{-6pt}
    \caption{Maze \#1}
\end{subfigure}
\hfill
\begin{subfigure}[b]{0.48\textwidth}
    \centering
    \includegraphics[width=0.9\textwidth]{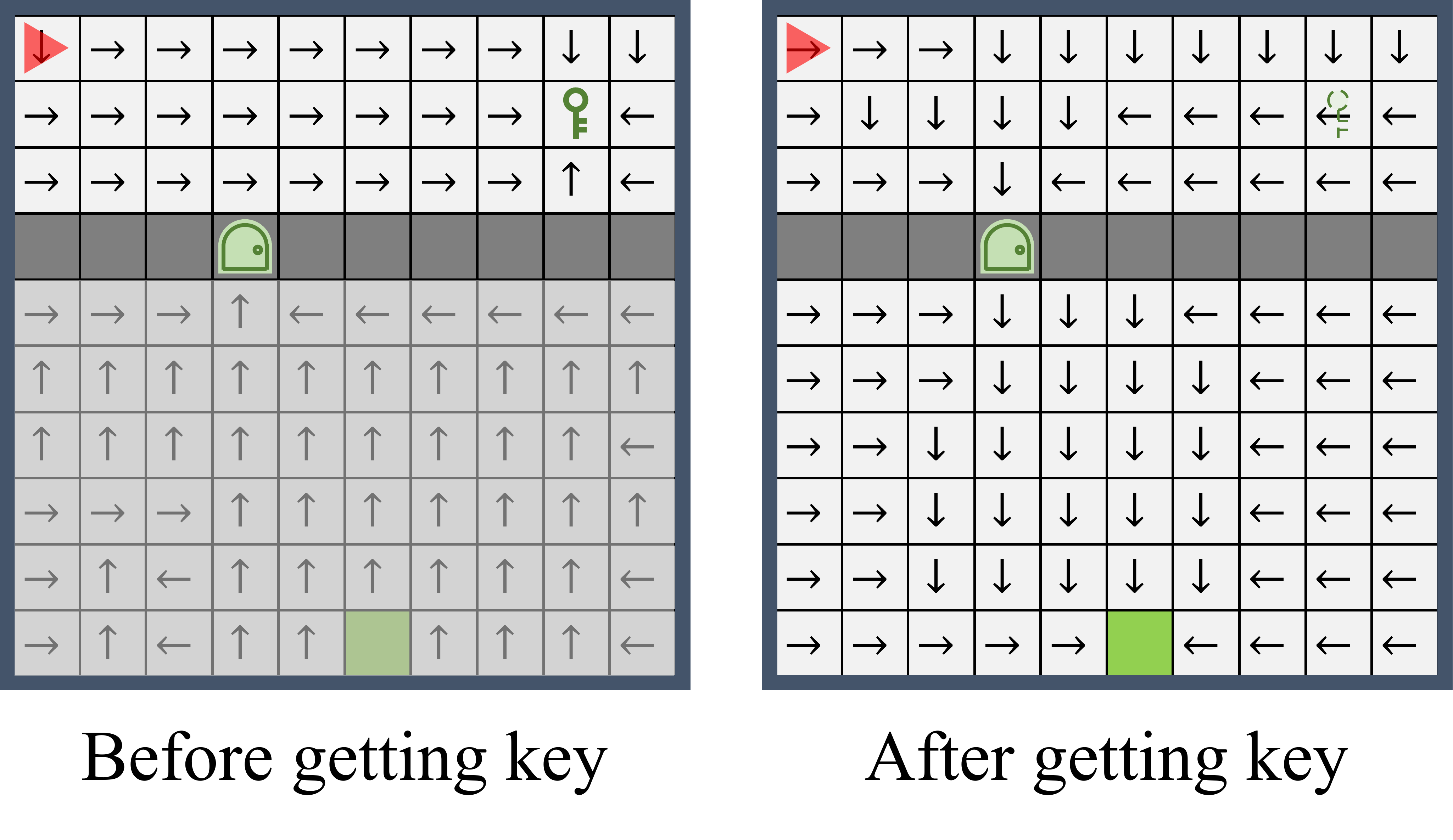}
    \vspace{-6pt}
    \caption{Maze \#2}
\end{subfigure}
\\
\vspace{10pt}
\begin{subfigure}[b]{0.48\textwidth}
    \centering
    \includegraphics[width=0.9\textwidth]{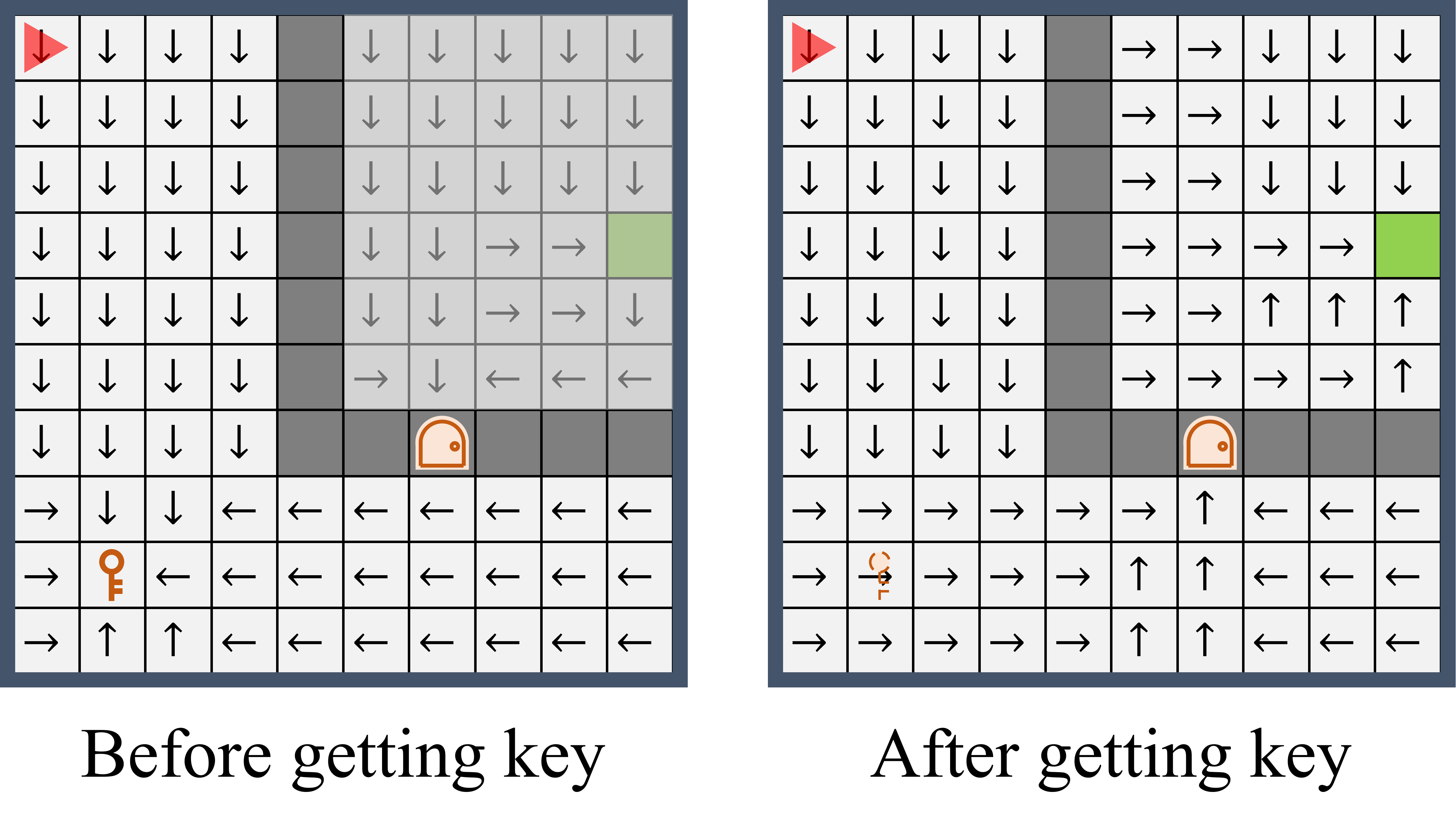}
    \vspace{-6pt}
    \caption{Maze \#3}
\end{subfigure}
\hfill
\begin{subfigure}[b]{0.48\textwidth}
    \centering
    \includegraphics[width=0.9\textwidth]{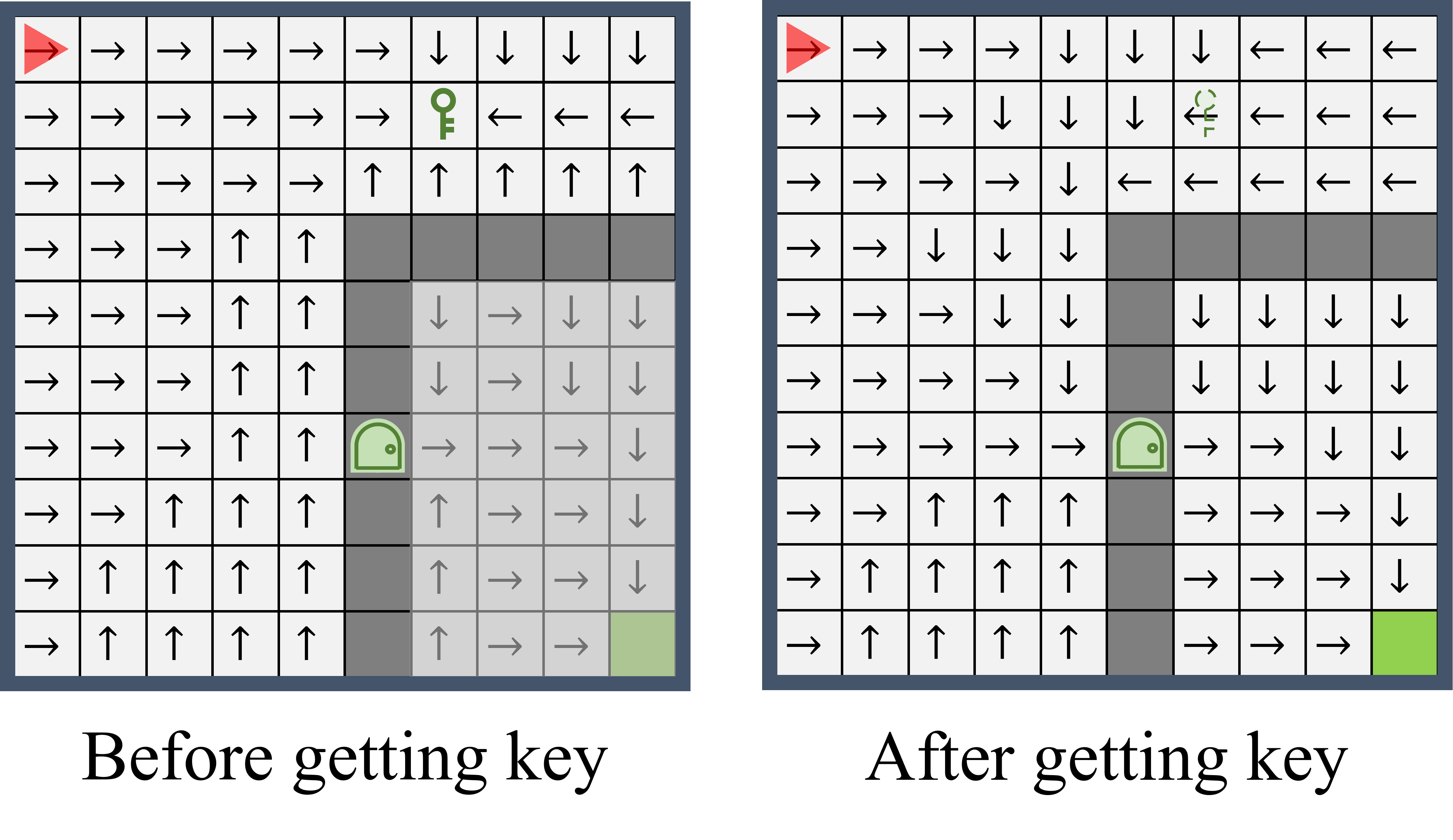}
    \vspace{-6pt}
    \caption{Maze \#4}
\end{subfigure}
\caption{The actions yielding the maximum knowledge rewards in the four \textit{2DMaze} tasks.}
\label{fig:learned-actions}
\end{figure}

In this section, we visualize the learned \textit{knowledge rewards} by the centralized reward agent $\mathcal{A}^{rwd}$ in the \textit{2DMaze} environment. After training on the four tasks in Section~\ref{sec:comp} of the paper, we let the CRA generate the knowledge rewards for each action in every state and visualize the action direction that yields the maximum rewards, $a^* = \argmax_{a}{{\pi^{rwd}}^*(s_i,a)}, s_i \sim S$, in Figure~\ref{fig:learned-actions}.

The shaded areas in the figures represent regions within the real task that the agent cannot reach, as it cannot access the space behind the door without picking up the key. However, we forced the agent into these areas for evaluation. Outside the shaded regions, we observe that the CRA successfully learned meaningful knowledge rewards. Before picking up the key, the agent received the highest reward in the corresponding state when moving towards the key. Similarly, after picking up the key, the agent received the highest reward when moving towards the door and the final target. This demonstrates that in scenarios where the original environmental rewards are sparse, these detailed knowledge rewards can effectively guide the agent to converge more quickly.

\end{document}